\documentclass[fleqn,10pt]{wlscirep}
\usepackage[utf8]{inputenc}
\usepackage[T1]{fontenc}
\usepackage{subfig}
\title{Secure \& Private Federated Neuroimaging}

\usepackage{amsmath,amssymb,amsfonts}

\author[1,+,*]{Dimitris Stripelis}
\author[1,+]{Umang Gupta}
\author[2,+]{Hamza Saleem}
\author[3]{Nikhil Dhinagar}
\author[1]{Tanmay Ghai}
\author[2]{Chrysovalantis Anastasiou}
\author[1]{Armaghan Asghar}
\author[1]{Greg Ver Steeg}
\author[1]{Srivatsan Ravi}
\author[2]{Muhammad Naveed}
\author[3]{Paul M. Thompson}
\author[1]{Jos\'{e} Luis Ambite}

\affil[1]{University of Southern California, Information Sciences Institute, Marina Del Rey, CA, 90292, USA}
\affil[2]{University of Southern California, Computer Science Department, Los Angeles, 90089, USA}
\affil[3]{University of Southern California, Imaging Genetics Center, Stevens Neuroimaging and Informatics Institute, Marina del Rey, CA, 90292 USA}

\affil[*]{stripeli@isi.edu}

\affil[+]{these authors contributed equally to this work}

\begin{abstract}

The amount of biomedical data continues to grow rapidly. However, collecting data from multiple sites for joint analysis remains challenging due to security, privacy, and regulatory concerns. 
To overcome this challenge, we use Federated Learning, which enables distributed training of neural network models over multiple data sources without sharing data. 
Each site trains the neural network over its {\em private data} for some time, then shares the neural network parameters (i.e., weights, gradients) with a Federation Controller, which in turn aggregates the local models, sends the resulting community model back to each site, and the process repeats. 
Our Federated Learning architecture, MetisFL, provides {\em strong security and privacy}. First, sample data never leaves a site. Second, neural network parameters are encrypted before transmission and the global neural model is computed under {\em fully-homomorphic encryption}. Finally, we use information-theoretic methods to {\em limit information leakage} from the neural model to prevent a ``curious'' site from performing  model inversion or membership attacks. 
We present a thorough evaluation of the performance of secure, private federated learning in {\em neuroimaging} tasks, including for predicting Alzheimer's disease and estimating BrainAGE from magnetic resonance imaging (MRI) studies, in challenging, heterogeneous federated environments where sites have different amounts of data and statistical distributions.

\end{abstract}

\begin{document}

\flushbottom
\maketitle
\thispagestyle{empty}

\section*{Introduction}
Deep learning and traditional machine learning methods are now widely applied across biomedical research~\cite{wainberg2018deep}. These methods have been particularly successful in medical imaging~\cite{suzuki2017overview}, including image reconstruction and enhancement~\cite{zhu2018image}, automated segmentation and labeling of key structures~\cite{dalca2019unsupervised}, computer-aided diagnosis~\cite{cho2021deep}, pathology detection~\cite{kofler2020brats}, disease subtyping~\cite{aksman2021pysustain,young2021ordinal}, and predictive analytics (e.g., modeling future recovery or decline)~\cite{ezzati2021predictive}. 

In neuroimaging, there is great progress in automated diagnostic classification and subtyping of diseases, such as Alzheimer's disease and Parkinson's disease, to assist in patient management and monitoring and to screen patients for eligibility for clinical trials. Some recent MRI-based classifiers have merged data from over 80,000 individuals for diagnostic classification~\cite{lu2021practical}. The performance of deep learning methods depends heavily on the availability of large amounts of training data. Unfortunately, data acquisition is expensive for many areas of biomedical research, such as neuroimaging. Therefore, any organization or research group can only collect limited data. 

To increase the amount of data and statistical power of analyses, research groups join together into consortia~\cite{thompson2020enigma}. However, the need to protect patient data makes data sharing very challenging. 
Regulatory frameworks, such as the \textit{Health Insurance Portability and Accountability Act} (HIPAA), require strict protection of health records and data collected for medical research. Privacy laws have spurred research into anonymization methods; for example, algorithms to remove facial information from MRI scans~\cite{bischoff2007technique,schimke2011quickshear,milchenko2013obscuring}. 
The inherent complexity and cost of enforcing security and privacy results in few large-scale data sharing efforts. Even when large consortia are established, they often only perform meta-analysis using traditional statistical methods instead of joint mega-analysis using deep learning methods. A paradigmatic example of large-scale meta-analysis is the ENIGMA Consortium~\cite{thompson2020enigma}.

Federated Learning \cite{mcmahan2017communication,yang2019federated,li2020federated} has emerged as a novel distributed machine learning paradigm that enables large-scale cross-institutional analysis without the need to move the data out of its original location. Federated Learning allows institutions to collaboratively train a machine learning model (e.g., a neural network) by aggregating the parameters (e.g., weights, gradients) of local models trained on local data. Since subject data is not shared and parameters can be protected through encryption, privacy concerns are ameliorated.
Federated Learning is being increasingly applied in biomedical and healthcare domains~\cite{lee2018privacy,sheller2018multi,silva2019federated,rieke2020future,silva2020fed}. 

\begin{figure}[htbp]
    \centering
    \includegraphics[width=0.8\textwidth]{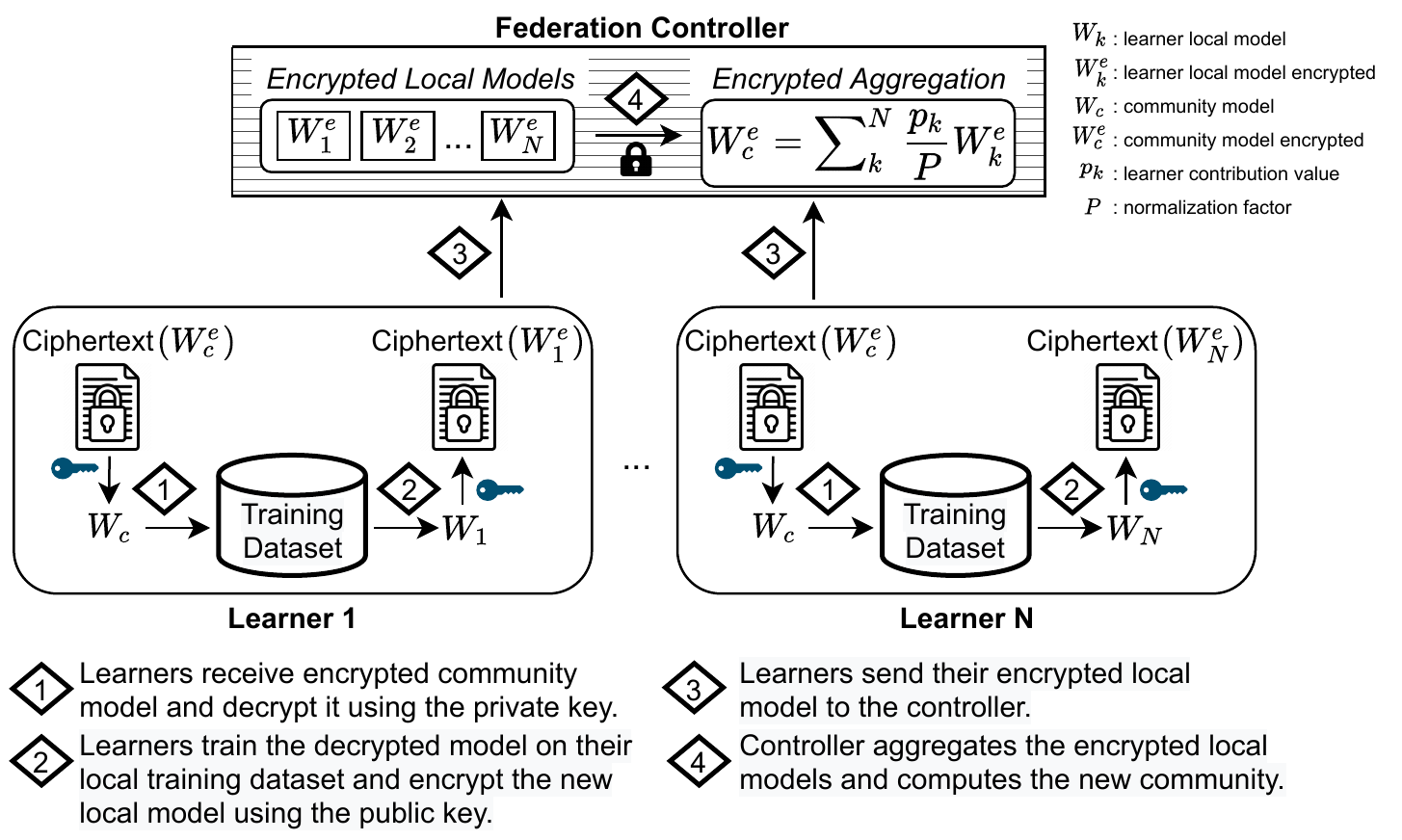}
    \caption{MetisFL, a Secure Federated Learning Architecture.}
    \label{fig:EncryptedFederatedSystemArchitecture}
\end{figure}

This paper demonstrates the potential of federated learning to accelerate and improve research outcomes through decentralized biomedical consortia. 
We conduct our analysis using MetisFL, our {\em Secure \& Private Federated Learning} system. 
Our design is modular, extensible, and supports a variety of federated training policies. 
The MetisFL architecture~\cite{stripelis2021secure} appears in Figure~\ref{fig:EncryptedFederatedSystemArchitecture}. 
Each site trains the neural network over its private data for some time, then shares the neural network parameters (i.e., weights, gradients) with a Federation Controller, which in turn aggregates the local models, sends the resulting community model back to each site, and
the process repeats. 
Federated training in MetisFL is secure. Data is never shared. Model parameters are transmitted through secure communication channels. Model parameters are encrypted, and the global model is computed under fully-homomorphic encryption (using CKKS \cite{ckks_paper}), so even if the controller was compromised, the global model cannot be attacked. Finally, we use information-theoretic methods to {\em limit information leakage} from the neural model to prevent a "curious" site within the federation from performing model inversion~\cite{geiping2020inverting, zhu2019deep} or membership inference~\cite{shokri2017,nasr2019} attacks.

We present a thorough evaluation of the performance of secure, private federated learning in neuroimaging tasks, including predicting Alzheimer’s disease and estimating BrainAGE from magnetic resonance imaging studies, in challenging, heterogeneous federated environments
where sites have different amounts of data and statistical distributions.
Specifically, we show that research consortia based on federated learning, without data sharing, can achieve comparable learning performance to centralized consortia, where data is shared into a single site. 
We show that that our homomorphic encryption methods are practical, having little runtime overhead over unencrypted training. 
We show defense mechanisms against attacks to federated neural models and the tradeoffs between security and learning performance. 
In summary, secure federated learning enables large, decentralized analysis of biomedical data, without the burdens of data sharing. Since the performance of deep learning models increases with the amount of data used for training, federated learning over research consortia promises improvements in disease diagnosis, prognosis, biomarker detection, and many other advances in biomedical research.

\section*{Results}

\paragraph{Federated learning can perform comparably to centralized analysis, and even outperform it in the likely scenario where more sites are willing to join a federation than those willing to share data.}
We evaluate centralized and federated learning on two challenging neuroimaging learning tasks: the Brain Age Gap Estimation (BrainAGE) regression task and the Alzheimer's Disease (AD) detection classification task, both on structural MRI inputs. 

For the {\bf BrainAGE} task, we select 10,446 MRI scans of healthy individuals (no neurological or psychiatric diagnoses) from the UK Biobank~\cite{miller2016multimodal}. We train and test a 3D Convolutional Neural Network (CNN) to predict BrainAGE, both on the centralized dataset, and on a federation of 8 sites under diverse data distributions. See Supplementary Figure~\ref{fig:3dcnn_model_definition} for the 3D-CNN model architecture.

\begin{figure}[htpb]
    \centering    
    \includegraphics[width=0.8\linewidth]{
    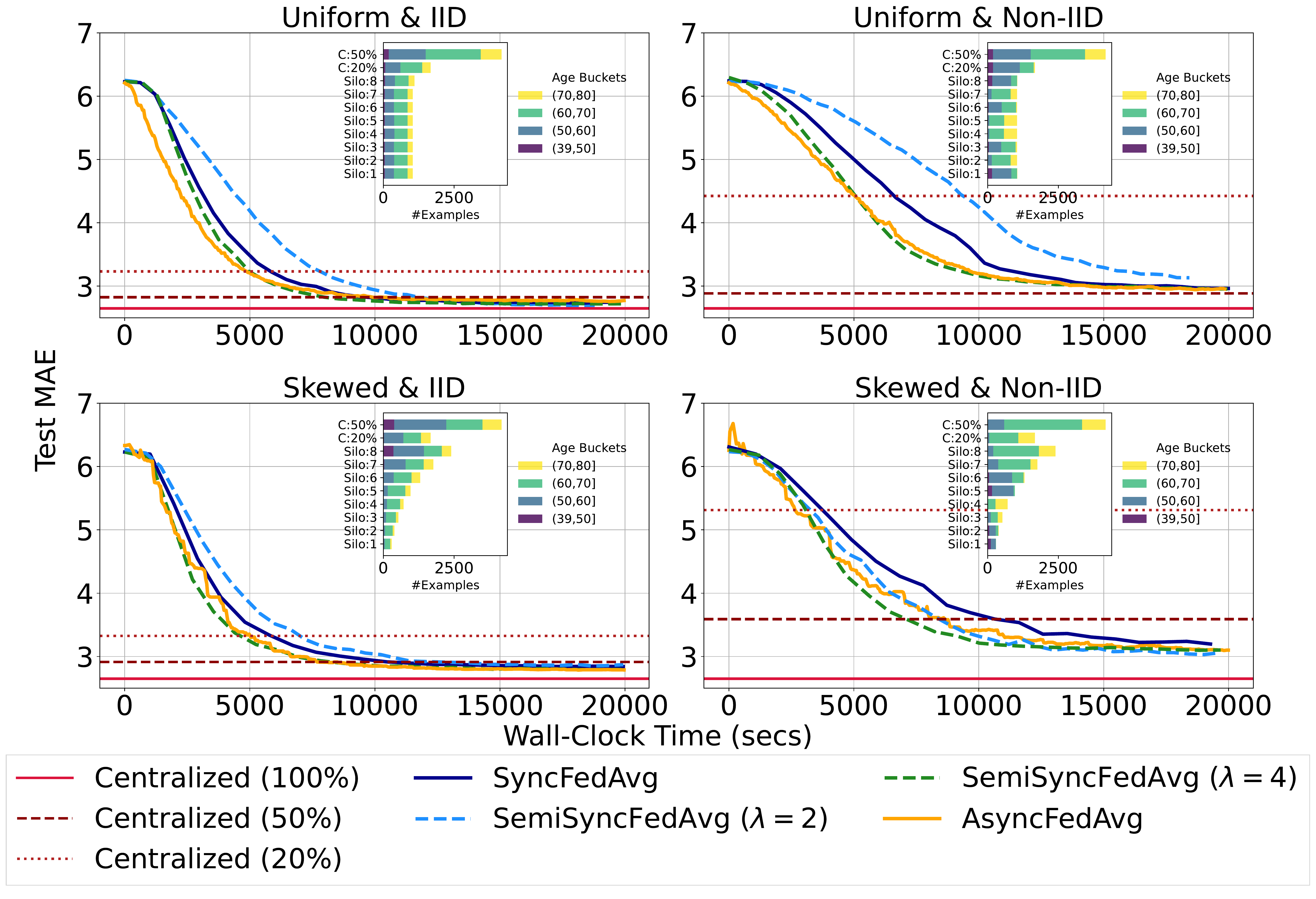}
    \caption{Federated and Centralized BrainAGE Models Comparison (Wall-Clock Time)}
    \label{fig:brainage_policies_convergence_wall_clock_time}
\end{figure}

Figure~\ref{fig:brainage_policies_convergence_wall_clock_time} shows the performance of the centralized and federated models in terms of wall-clock time execution. Supplementary Figure~\ref{fig:BrainAge3D_PoliciesConvergence_CommunicationCost} shows model convergence based on communication cost. We tested a diverse set of federated environments with different amounts of data per site (Uniform: an equal number of training samples per site; Skewed: a decreasing amount of training samples for each site), and different data distributions (IID: Independent and Identically Distributed, where the local data distribution of each site is similar to the global distribution, and Non-IID, where it differs). The data distributions appear as insets in Figure~\ref{fig:brainage_policies_convergence_wall_clock_time}. 
See Supplementary Materials and Figures~\ref{fig:UKBB_Age_Distributions_ALL},~\ref{fig:UKBB_Centralized_Age_Distributions}, and~\ref{fig:UKBB_Federation_Age_Distributions} for details on the data distributions. 
The horizontal red lines show the performance of centralized models, while the other lines show the convergence of federated training under different policies. All models are evaluated against the same test set.

In the Uniform and IID environment, federated training achieves the same performance as centralized training, over the complete 10K MRI dataset. In the harder Skewed or Non-IID environments, there is a (small) gap between centralized and federated analysis when the same amount of data is available to both approaches. 
However, the promise of federated learning is that federated consortia, which do not require data sharing, can enroll more sites than centralized consortia, which require data sharing. 
Therefore, we also show the performance of centralized systems that can only obtain a fraction of the data of the federated system, specifically 50\% or 20\% of the data; in other words, the federation is composed of sites that in total have double, or five times the amount of data that can be shared centrally. In the BrainAGE task, regardless of the data distributions, federated training always significantly outperforms a centralized system when the federation reaches five times more data, which is feasible; and outperforms or matches centralized training when the federation reaches double the data, which is a reasonable assumption.
We expect that without the burden of data sharing much larger federated consortia can be formed and yield better analyses.

For the {\bf Alzheimer's disease detection} task, we analyze three well-known studies: the Alzheimer’s Disease Neuroimaging Initiative (ADNI)~\cite{mueller2005alzheimer}, with its three phases ADNI\;1, ADNI\;2 and ADNI\;3; the Open Access Series of Imaging Studies (OASIS)~\cite{LaMontagne2019.12.13.19014902}; and   the Australian Imaging, Biomarkers \& Lifestyle Flagship Study of Ageing (AIBL)~\cite{fowler2021fifteen}. 
These studies contain T1-weighted brain MRIs taken from patients with different degrees of dementia and healthy subjects acting as controls. 
For our work, we only use images from control subjects and patients diagnosed with Alzheimer's disease. 
These studies are longitudinal. To obtain unbiased performance estimates, all the samples for a given subject appear either in training or the test set. 
Table~\ref{tbl:AD_data_distribution} shows the number of training and testing samples from each study and the target labels. 

\begin{table}[htbp]
\centering
\begin{tabular}{@{}lcccc@{}}
\toprule
& \multicolumn{2}{c}{Training Set} & \multicolumn{2}{c}{Test Set} \\
\cmidrule(lr){2-3}
\cmidrule(lr){4-5}
\multicolumn{1}{c}{\textbf{Cohort}} & Alzheimer's & Controls & Alzheimer's & Controls \\
\cmidrule(){1-1}
\cmidrule(lr){2-2}
\cmidrule(lr){3-3}
\cmidrule(lr){4-4}
\cmidrule(l){5-5}
OASIS-3 & 315 & 1113 & 68 & 209 \\
AIBL & 113 & 642 & 28 & 160 \\
ADNI-1 & 759 & 940 & 299 & 256 \\
ADNI-2 & 534 & 1137 & 185 & 399 \\
ADNI-3 & 90 & 388 & 26 & 118 \\
\textit{Total} & \textit{1,811} & \textit{5,220} & \textit{606} & \textit{1,142} \\
\bottomrule
\end{tabular}
\caption{AD: Train/test splits per cohort and target label.}
\label{tbl:AD_data_distribution}
\end{table}

We compare the performance of models trained on the dataset of a single cohort, that is, a centralized model for each of OASIS-3, AIBL, and ADNI-\{1,2,3\}, to models trained on a federation with an increasing number of sites/cohorts, that is, a federation of 3 sites, one with each of the 3 ADNI phases (ADNI-\{1,2,3\}), a 4-site federation (ADNI-\{1,2,3\} + AIBL), and a 5-site federation (ADNI-\{1,2,3\} + AIBL + OASIS-3). 
All environments train the same 3D-CNN neural architecture (shown in Supplementary Figure~\ref{fig:3dcnn_model_definition}). 
Table~\ref{tbl:AD_results} shows that model obtained by the federation outperforms models trained at any single site, and has comparable AUC ROC with the centralized model trained over all the data. 
The AUC ROC provides a robust measure of classifier performance, since it does not depend on a specific threshold.

There is a larger difference in precision and recall values, reflecting a sensitivity to the classification threshold likely due to the class imbalance. 
The greater the number of cohorts participating in the federation, the better the predictive performance of the learned model is.

\begin{table*}[htbp]
\centering
\small
\setlength{\tabcolsep}{4.5pt}
\begin{tabular}{@{}lcccccc@{}}
\toprule
\multicolumn{1}{c}{\textbf{Model}} & \textbf{AUC ROC} & \textbf{Accuracy} & \textbf{Precision} & \textbf{Recall} & \textbf{F1} \\
\cmidrule(){1-1}
\cmidrule(lr){2-2}
\cmidrule(lr){3-3}
\cmidrule(lr){4-4}
\cmidrule(lr){5-5}
\cmidrule(lr){6-6}
(C) OASIS-3 & 0.6106 $\pm$ 0.0032 & 0.6522 $\pm$ 0.0004 & 0.4631 $\pm$ 0.0092 & 0.0283 $\pm$ 0.0016 & 0.0535 $\pm$ 0.0028 \\
(C) AIBL & 0.5321 $\pm$ 0.0051 & 0.6082 $\pm$ 0.0301 & 0.3661 $\pm$ 0.0074 & 0.0886 $\pm$ 0.0161 & 0.1367 $\pm$ 0.0159 \\
(C) ADNI-\{1,2,3\} & 0.7441 $\pm$ 0.0395 & 0.6421 $\pm$ 0.0542 & 0.4763 $\pm$ 0.0659 & 0.5858 $\pm$ 0.1617 & 0.5251 $\pm$ 0.1093 \\
\textbf{(C) ADNI-\{1,2,3\} + AIBL + OASIS-3} & \textbf{0.8789 $\pm$ 0.0004} & 0.7723 $\pm$ 0.0072  & 0.9105 $\pm$ 0.0019 &  0.3775 $\pm$ 0.0268 & 0.5362 $\pm$ 0.0279 \\
(F) ADNI-\{1,2,3\} & 0.7683 $\pm$ 0.0013 & 0.6898 $\pm$ 0.0021 & 0.5378 $\pm$ 0.0019 & 0.7451 $\pm$ 0.0086 & 0.6257 $\pm$ 0.0033 \\
(F) ADNI-\{1,2,3\} + AIBL & 0.8339 $\pm$ 0.0039 & 0.7493 $\pm$ 0.0018 & 0.6162 $\pm$ 0.0034 & 0.7338 $\pm$ 0.0061 & 0.6745 $\pm$ 0.0033 \\
\textbf{(F) ADNI-\{1,2,3\} + AIBL + OASIS-3} & \textbf{0.8756 $\pm$ 0.0029} & 0.8282 $\pm$ 0.0005 & 0.7715 $\pm$ 0.0066 & 0.7148 $\pm$ 0.0112 & 0.7437 $\pm$ 0.0021 \\
\bottomrule
\end{tabular}
\caption{Alzheimer's Disease Prediction. Test results on a global stratified test dataset (5 sites), for each dataset by itself in a centralized environment (OASIS-3, AIBL); a federation of 3 sites (ADNI-\{1,2,3\}), 4 sites (ADNI-\{1,2,3\} + AIBL), and all 5 sites (ADNI-\{1,2,3\} + AIBL + OASIS-3).  With (C) and (F) we denote the centralized and federated environments, respectively. In the federated environments, each dataset is located at a different site. Centralized environments are trained over all the corresponding datasets. The evaluation is conducted over three different runs.}
\label{tbl:AD_results}
\end{table*}

\paragraph{Fully Homomorphic Encryption can efficiently protect Federated Learning against attackers outside of the federation.} 
Neural models can memorize training data and are susceptible to model inversion attacks~\cite{geiping2020inverting, zhu2019deep} or membership inference attacks~\cite{shokri2017,nasr2019}. Therefore, if the sites shared unprotected models with a compromised federation controller, or the models were captured in transit, an attacker may obtain private information. 
To prevent such attacks against the local neural models, we use fully homomorphic encryption (FHE), specifically the CKKS scheme~\cite{ckks_paper}. The sites encrypt their local models before transmission and the federated controller aggregates the models in an encrypted space. The optimizations to FHE in our MetisFL architecture provide low runtime overhead compared to unencrypted training.

\begin{figure}[htpb]
    \centering    
    \includegraphics[width=0.8\linewidth]{
    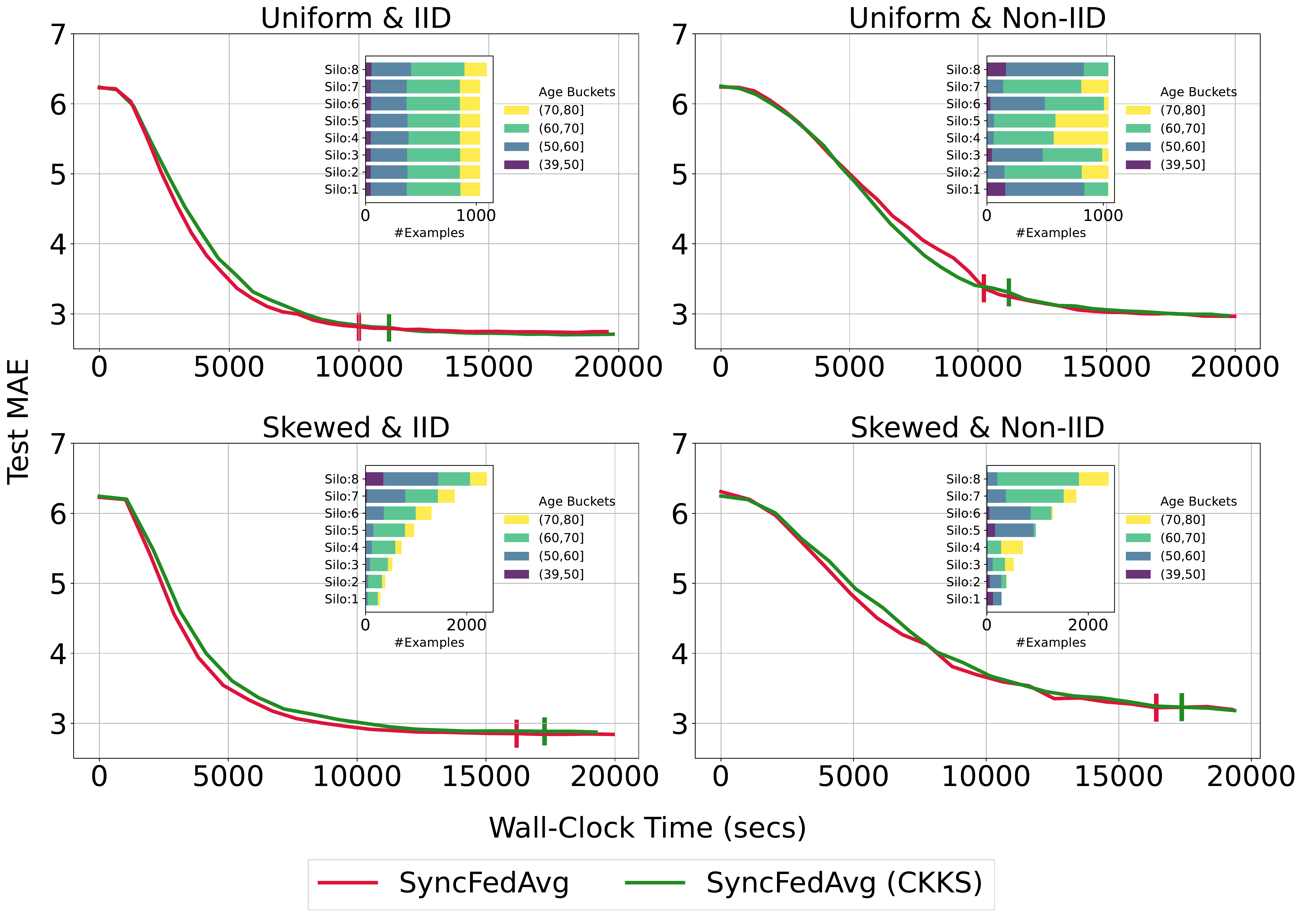}
    \caption{Federated Learning (SyncFedAvg) with and without CKKS homomorphic encryption on the BrainAge 3D-CNN.
The vertical marker represents the training time it takes for each approach to complete 20 federation rounds.}
    \label{fig:BrainAgeEvaluation_CKKS}
\end{figure}

We evaluate the learning performance of CKKS FHE on the BrainAGE prediction task over four federated learning environments using a 3D-CNN model with 3 million parameters (Supplementary Figure~\ref{fig:3dcnn_model_definition}). Figure~\ref{fig:BrainAgeEvaluation_CKKS} shows the execution (Wall-Clock) time of synchronous federated average (SyncFedAvg) with and without encryption (Supplementary Figure~\ref{fig:BrainAge3D_FHE_PoliciesConvergence_CommunicationCost} shows model convergence based on communication cost). Learning performance is almost identical, at a small ($\sim$7\%) additional training time cost. Therefore, our MetisFL system can provide strong privacy guarantees in practice.

\paragraph{Federated learning is vulnerable to insider attacks.} 

Our architecture uses homomorphic encryption for secure communication and aggregation of parameters. Thus, the system is protected against attacks from outside the federation or a compromised controller. However, each learner needs to decrypt the community model for local training.
Therefore, a {\em curious} site inside the federation may attempt a model inversion attack~\cite{geiping2020inverting, zhu2019deep}, or a membership inference attack~\cite{shokri2017,nasr2019} against the community model (the local models of the other learners are protected through encryption).
Model inversion attacks against the global federation model are impractical in realistic
settings, since learners train the global model locally for a large number of local iterations, and the local models are aggregated. Therefore the risk of leaking any identifiable information from the global model is limited~\cite{hatamizadeh2022gradient}.
In contrast, membership inference attacks are very successful. A site can use private local subject data to probe the federated community model and discover if such data was used for training.  
Discovering that the MRI scan of a particular subject was used for training the model can reveal information about a person's medical history or participation in some sensitive medical study~\cite{gupta2021membership}. 

\begin{figure}[htpb]
    \centering
    \includegraphics[width=0.5\linewidth]{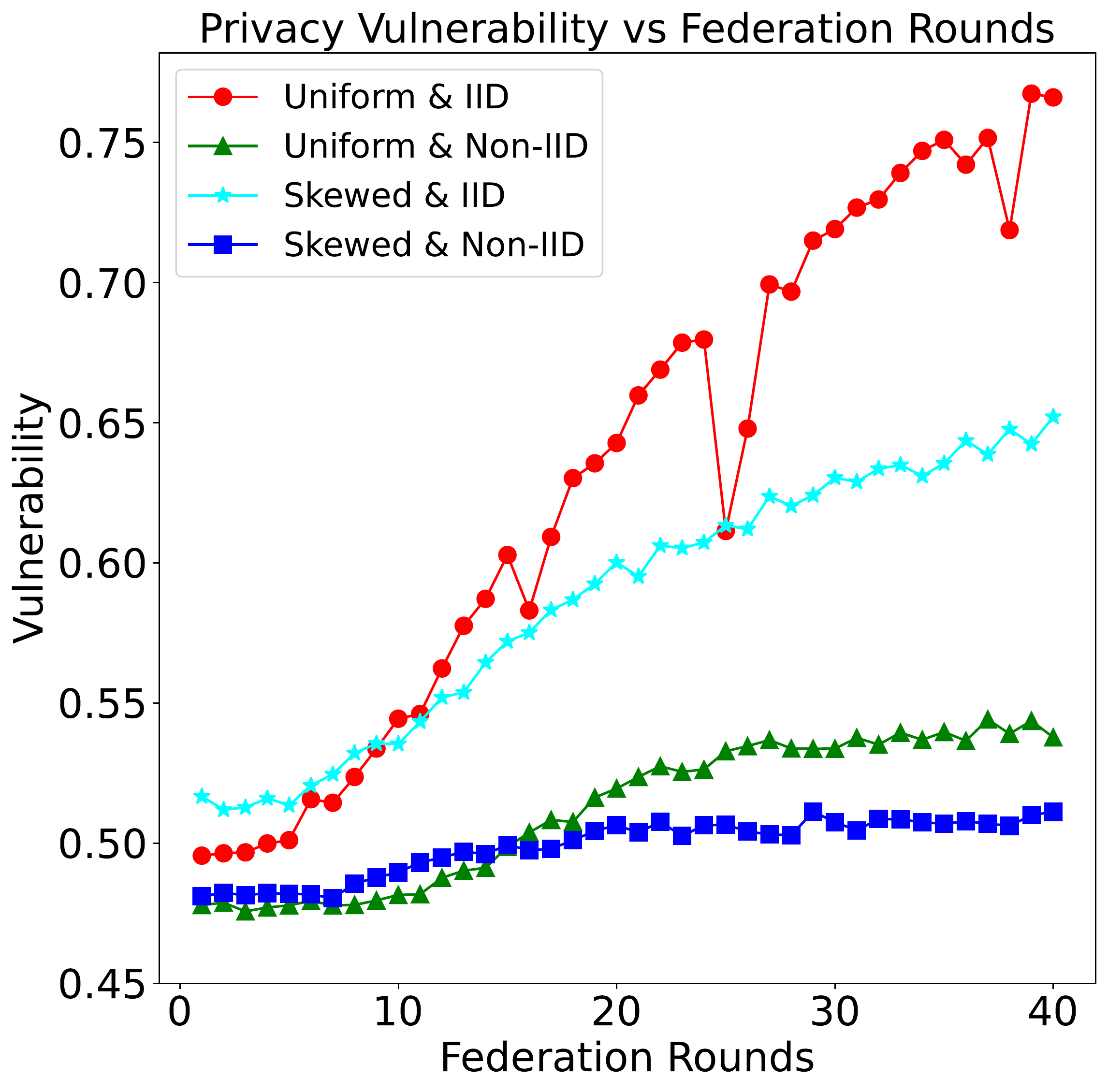}
    \caption{Privacy vulnerability increases with federation rounds. Vulnerability is the average accuracy of distinguishing train samples vs. unseen samples across sites.}%
    \label{fig:vulnerability_per_round}
\end{figure}

Figure~\ref{fig:vulnerability_per_round} shows the increasing vulnerability of community models at each round. 
We use the features and architecture in~\cite{gupta2021membership} to conduct the membership inference attacks. 
We measure vulnerability as the average accuracy of detecting that an MRI was used for training over 56 different datasets (see Methods). 
As training progresses, the neural network learns more information about the samples, and it becomes easier to identify MRIs participating in training. Notably, in the Uniform \& IID environment attack success reaches 80\%. 
Previous works~\cite{gupta2021membership, yeom2018privacy, truex2018towards, salem2019ml} have also found a strong correlation between overfitting and vulnerability to these attacks. We see that data distribution across silos may impact vulnerability. Models trained over more homogenous (IID) data distributions across silos are more vulnerable than heterogenous data distributions (Non-IID), as Non-IID distributions may implicitly regularize and reduce overfitting. Overall, vulnerability increases during training, suggesting a trade-off between learning performance and privacy risk.

\paragraph{Federated training with gradient noise protects against membership attacks from insiders.} 
The success of privacy attacks is often attributed to the ability of learning algorithms to memorize information about a single sample~\cite{jha2020extension}. Therefore, defending against data privacy leakages involves limiting the information the learning algorithm may extract about each sample or limiting information in the neural network's weights. We explore approaches to limit the vulnerability to membership inference attacks: differential private training via DP-SGD~\cite{abadi2016deep} and SGD with non-unique gradients~\cite{gupta2021membership}. 

\begin{figure}[htpb]
\centering
    \subfloat[Uniform \& IID]{
        \centering
        \includegraphics[width=0.4\linewidth]{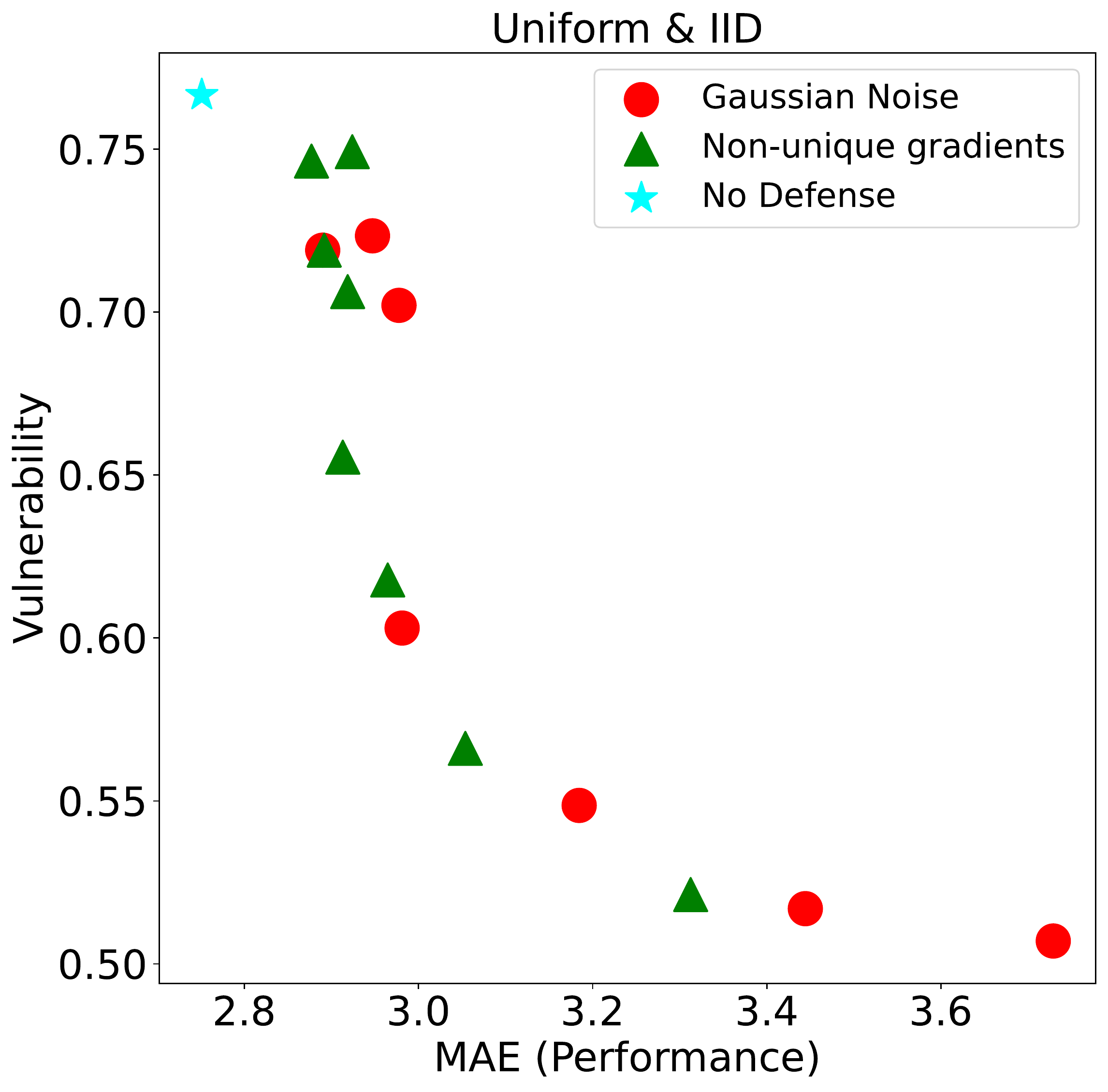}
        \label{fig:privacy_trade_off_uniform_iid}
    }
    \subfloat[Uniform \& Non-IID]{
        \centering
        \includegraphics[width=0.4\linewidth]{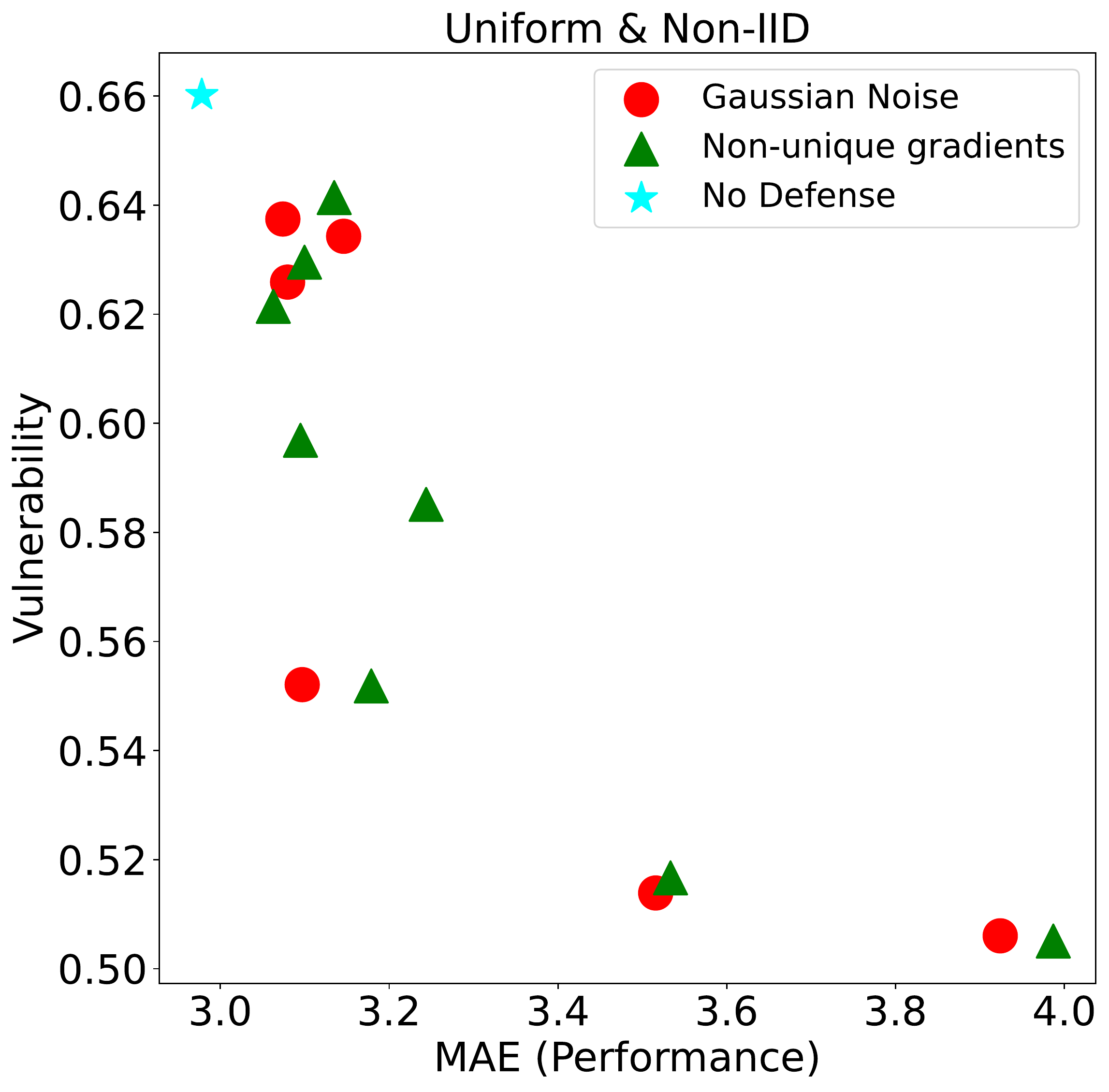}
        \label{fig:privacy_trade_off_uniform_noniid}
    }
    \caption{Vulnerability vs. Performance trade-off when training sites with Differential Privacy (Gaussian Noise) and Non-unique gradients approach. Lower vulnerability and lower MAE are desired, i.e., points towards the bottom left are better.}
    \label{fig:privacy_trade_off}
\end{figure}

Figure~\ref{fig:privacy_trade_off} shows the privacy and learning performance trade-off when sites are trained with small-magnitude Gaussian noise and our non-unique gradient approach. Both approaches can reduce the vulnerability of the global model to privacy attacks. Although the magnitude of Gaussian noise is much smaller than the theoretically-required differential privacy guarantees, it effectively reduces membership inference attacks. 

\section*{Discussion}
Dementia affects more than 50 million people worldwide, and this number could exceed 152 million by 2050~\cite{patterson2018world}, with Alzheimer's disease (AD) being the leading cause. Recently, deep learning has been applied to identify AD from structural brain MRI scans~\cite{lu2021practical,10.1117/12.2606297,abdulazeem2021cnn,fu2021automated}. Similar in spirit, the Brain Age Gap Estimation (BrainAGE) task is another pathway toward assessing the acceleration or deceleration of an individual's brain aging through structural MRI scans. The difference between the true chronological age and the predicted age of the brain is considered an important biomarker for early detection of age-associated neurodegenerative and neuropsychiatric diseases~\cite{franke2019ten,wood2022accurate}, such as cognitive impairements~\cite{cole2015prediction}, schizophrenia~\cite{koutsouleris2014accelerated}, chronic pain~\cite{kuchinad2007accelerated}.

Recently, deep learning methods based on RNN~\cite{lam2020accurate,jonsson2019brain} and CNN~\cite{dinsdale2021learning,cole2017predicting,peng2021accurate,gupta2021improved} architectures have demonstrated accurate brain age predictions. We extended these methods to train a 3D-CNN model over centralized and federated environments with highly heterogeneous data distributions~\cite{stripelis2021scaling,stripelis2022semi} for the BrainAGE and AD prediction tasks. We show that federated learning can achieve comparable performance to centralized training. We posit that, ultimately, federated consortia will allow to analyze much greater quantities of data since they sidestep many of the challenges of centralized data sharing. 

Even though federated learning avoids sharing datasets, the exchanged parameters may reveal private information. Various works have highlighted this through practical privacy attacks such as model inversion attacks~\cite{geiping2020inverting, zhu2019deep} and membership inference~\cite{shokri2017,nasr2019}, in both centralized and federated settings. Researchers have focused on reducing overfitting~\cite{truex2018towards, salem2019ml}, information in activations and weights~\cite{jha2020extension}, or using differential private mechanisms to alleviate such privacy concerns. Learning under differential privacy is particularly attractive as it comes with theoretically solid worst-case guarantees. 
However, these works assume different threat models, which may relax the problem. For example, \cite{wei2020federated} assumes that the server can be trusted, whereas \cite{noble2021differentially, zhao2020local} consider a stricter threat model, considering the server as honest-but-curious similar to us. Rather than using a theoretical upper-bound measure of privacy, we focus on a more practical measure (i.e., membership inference attacks). We study membership inference attacks in our framework using the white-box experimental setup from Gupta et al.~\cite{gupta2021membership}. We assume that the attacker has access to the model, some samples that participated in the training, and some samples the attacker is curious about. We show that federated training with noise protects the models from attacks and the trade-off between protection and learning performance.

We use Homomorphic Encryption (HE)~\cite{HomomorphicEncryptionSecurityStandard} to ensure that models are protected from attacks from outside the federation. HE is a public-key encryption scheme~\cite{Sako2011} that enables certain computations (e.g., additions, multiplications) to be directly performed over encrypted data without decrypting them first. This distinct computational property renders HE a valuable cryptographic scheme for preserving data privacy in distributed settings, as untrusted parties can be tasked with performing operations over ciphertexts.
In our setting, we consider an honest-but-curious threat model and assume the participating parties do not collude with each other. To ensure secure model communication and aggregation, we use CKKS, a fully homomorphic encryption (FHE) scheme~\cite{ckks_paper}. Compared to the Paillier scheme used in past works~\cite{DBLP:journals/corr/abs-1812-03224, batchcrypt_paper}, CKKS supports arithmetic operations over real and complex data and is orders of magnitudes faster and can support an unbounded amount of additions and multiplications over encrypted data. While some previous works~\cite{DBLP:journals/corr/abs-1812-03224, ma-mkckks} may leak the global model to the controller, our protocol ensures that no global model leakage occurs at the controller. 

In summary, we presented the MetisFL Secure and Private Federated Learning system, a practical, extensible architecture supporting a variety of communication protocols with strong privacy and security mechanisms. Specifically, MetisFL provides protection against attacks from outside the federation through efficient homomorphic encryption, and against insider attacks by adding small, targeted noise during federated training. 
We demonstrated its efficacy on Neuroimaging tasks, BrainAGE estimation, and Alzeihmer's Disease prediction, over challenging statistically heterogeneous environments. We showed that federated learning can achieve the same learning performance as centralized learning in realistic environments. In hard heterogeneous environments, a small performance gap remains. 
We expect centralized consortia, which require data sharing, will include fewer sites than federated consortia, which do not share data. We posit that the larger consortia using federated learning promise to yield better analysis and greater advances in biomedical research.

\section*{Methods}

\paragraph{Federated Learning.} 
A federated learning environment consists of $N$ sites (learners, clients) that jointly train a machine learning model, often a neural network. The goal is to find the model parameters $w^*$ that optimize the global objective function $F(w): w^*=\underset{w}{\mathrm{argmin}}\;F(w) = \sum_{k=1}^{N}\frac{p_k}{P}F_k(w)$,
where $F_k(w)$ denotes the local objective function of learner $k \in N$ optimized over its local training dataset $D_k$. 
The global model, $F(w)$, is computed as a weighted average of the learners' local models, $P=\sum_k^N|p_k|$. 

A typical policy~\cite{mcmahan2017communication}, which we follow in this paper, is to weigh each local model based on the number of training examples it was trained on, i.e., $p_k$=$|D_k|$, $P=\sum_k^N|D_k|$, though other methods are possible \cite{yang2019federated,stripelis2020:dvw,stripelis2022:corrupted}.
Typically, each learner uses Stochastic Gradient Descent (SGD) to optimize its local objective on its local dataset. At a given synchronization point, each learner shares its local model parameters with the Federation Controller, which aggregates the local models (e.g., using weighted average) to compute a global (or community) model, sends it back to the learners, and the process repeats. Each such cycle is called a federation round. 
This iterative process was first introduced in the seminal work of~\cite{mcmahan2017communication} and it is termed as \textit{FedAvg}.

More recent works~\cite{wang2021field,DBLP:conf/iclr/ReddiCZGRKKM21,hsu2019measuring} have proposed a more general federated learning optimization framework where the optimization problem is split into global (server-side) and local (client-side). Global targets to optimize the merging rule of the learners' models updates into the global model and local targets to optimize learners' local training.
During training, a learner only shares its local model parameters with the Federation Controller. Each local dataset remains private.

\paragraph{Predicting BrainAGE.} In our experiments we use the same 3D-CNN architecture as in Stripelis et al.~\cite{stripelis2021scaling}, but without the dropout layer. This slight modification improved the learning performance on the BrainAGE task for both the centralized and federated models across all environments.
The training and testing datasets follow~\cite{stripelis2021scaling,stripelis2022semi}. We selected 10,446 scans (out of 16,356) from the UKBB~\cite{miller2016multimodal} dataset with no indication of neurological pathology, and no psychiatric diagnosis as defined by the ICD-10 criteria. All scans were evaluated with a manual quality control procedure and processed using a standard preprocessing pipeline with non-parametric intensity normalization for bias field correction and brain extraction using FreeSurfer and linear registration to a (2 mm)$^3$ UKBB minimum deformation template using FSL FLIRT. The final dimension of each scan was 91x109x91. Of the 10,446 scans, we used 8356 for training and 2090 for testing.
We generated four computationally and statistically heterogeneous federated learning environments comprising eight sites (learners). Computationally, the first four learners were equipped with NVIDIA GeForce GTX 1080 Ti GPUs, while the last four had (faster) Quadro RTX 8000 GPUs. 

For data amounts, we considered both {\em Uniform}, an equal number of training samples per learner, and {\em Skewed}, decreasing amount of training samples for each learner. For data distributions, we considered both \textit{Independent and Identically Distributed (IID)}, the local data distribution of each learner contains scans with the same distribution as the global age distribution, and {\em Non-IID}, different age distributions. 
We tested synchronous (SyncFedAvg), asynchronous (AsyncFedAvg), and semi-synchronous (SemiSyncFedAvg)~\cite{stripelis2022semi} federated training policies, all using the training data size as a weighting rule. 
We evaluate the performance of all policies against the same holdout test dataset to estimate the mean absolute error (MAE) between individuals' true chronological brain age and predicted age.

Each site (learner) is trained using Stochastic Gradient Descent (SGD) with a learning rate of 5x$10^{-5}$ and a batch size of 1. For SyncFedAvg and AsyncFedAvg, local training was performed over for 4 epochs. For SemiSyncFedAvg we evaluated synchronization periods $\lambda$ equivalent to the time it took the slowest learner to complete 4 or 2 epochs. 
AsyncFedAvg uses the caching method we introduced in Stripelis et al.~\cite{stripelis2022semi}, significantly improving performance. The controller stores each learner's most recently committed local model in a cache. When a learner issues an update request, the controller replaces its previously cached model and computes the new global model by performing a weighted average using all cached models.

Figure~\ref{fig:brainage_policies_convergence_wall_clock_time} compares the learning convergence of the training policies based on elapsed execution (wall-clock) time; we also provide a comparison in terms of communication cost in the supplemental material. 
In IID environments, both for Uniform and Skewed data amounts, federated training achieves comparable learning performance (MAE) to centralized training. 
The asynchronous protocol, AsyncFedAvg, is competitive in task performance but requires significantly more communication. SemiSync has fast convergence with low communication costs.

With Centralized-20\%, 50\%, and 100\%, we aim to emulate small, medium, and large research consortia that have established data sharing agreements to share their local data with a central authority for further analysis. 
To generate the centralized datasets for every environment, we start assigning data samples from the silo owning the majority of data samples (i.e., silo:8) until we reach 20\% and 50\% of the total data. 
 That is, we assume that the first few sites in the consortia decide to share data. 
All models (centralized and federated) are evaluated on the same test dataset.

\paragraph{Predicting Alzheimer's Disease.} In our evaluation, we studied 3 prominent AD studies: the 3 phases of ADNI~\cite{mueller2005alzheimer}, OASIS-3~\cite{LaMontagne2019.12.13.19014902}, and AIBL~\cite{fowler2021fifteen}. 
Images were preprocessed following the pipeline in Dhinagar et al.~\cite{10.1117/12.2606297}. First, images were reoriented using fslreorient2std (FSL v6.0.1), so to match the orientation of standard template images. Then, brain extraction was performed: skull parts in the image were removed using the HD-BET CPU implementation, and grey- and white-matter masks were extracted using FSL-FAST (FSL v6.0.1 Automated Segmentation Tool). An intensity normalization step (N4 bias field correction) using ANTs (v2.2) followed. Next, linear registration to a UK Biobank minumum deformation template was obtained by using the FSL-FLIRT (FSL v6.0.1 Linear Image Registration Tool) with 6 degrees of freedom. Finally, an isometric voxel resampling to 2mm was applied using the ANTs ResampleImage tool (v2.2). The actual size of the images after the preprocessing were volumes of 91x109x91 voxels.

We trained a 3D-CNN neural model over a federation of 3 (ADNI phases), 4 (ADNI phases + OASIS), and 5 learners (ADNI phases + OASIS + AIBL). Table~\ref{tbl:AD_results} shows the performance of the federated and the centralized models. The federated models were trained using the synchronous protocol (i.e., SyncFedAvg) for 40 federation rounds with each learner training locally for 4 epochs in-between rounds. The centralized models were trained for 100 epochs. All models were trained using Adam with Weight Decay with a learning rate of 1e-5 and weight decay of 1e-4. All experiments were run 3 times and the results show the average and standard deviation of the metrics.

\begin{figure*}[htpb]
    \centering
    \includegraphics[width=0.9\linewidth]{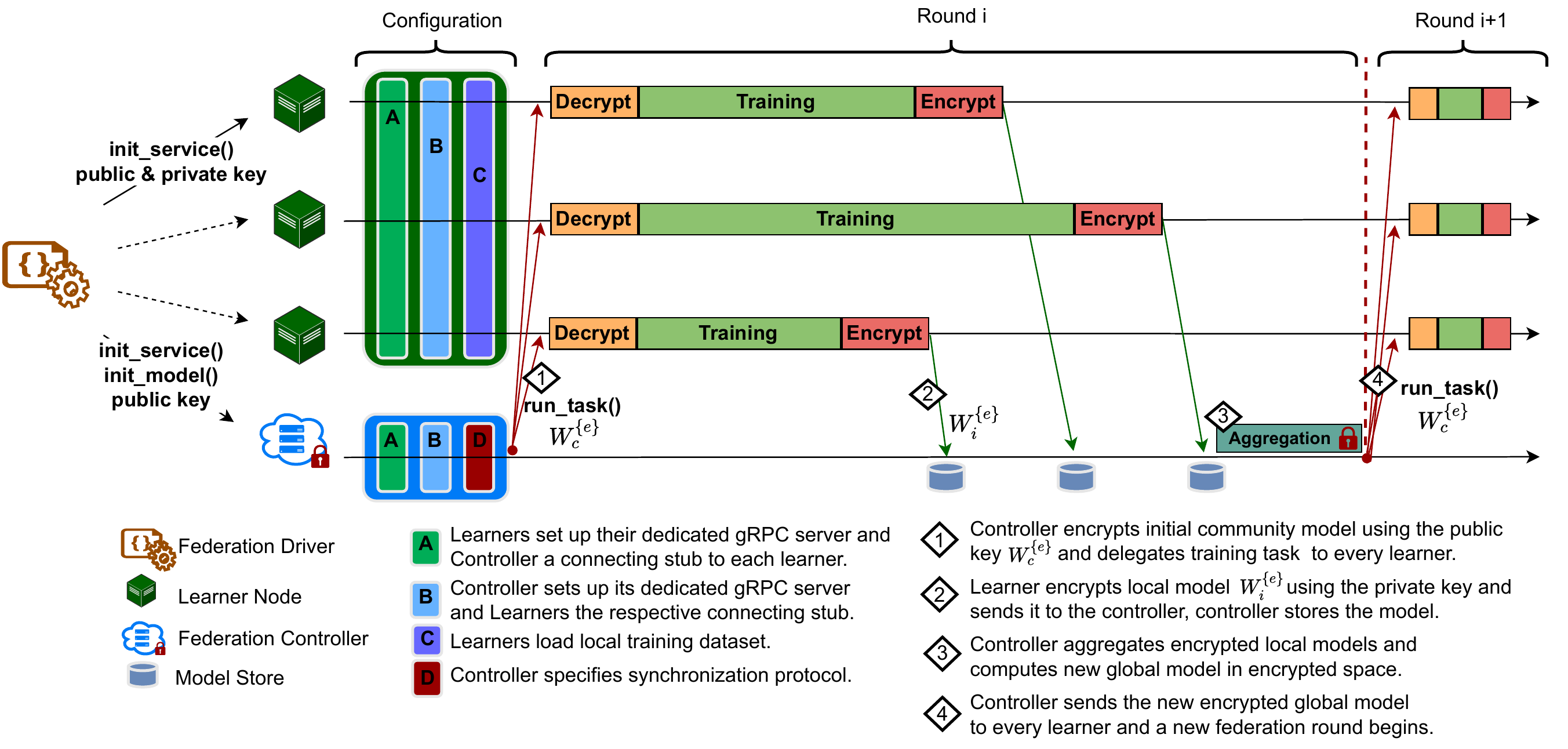}
    \caption{MetisFL homomorphic encryption training pipeline using the synchronous communication protocol. After the federation configuration, controller sends the original encrypted model to each learner, learners decrypt and train the received global model, then encrypt and send the new local model to controller and the controller aggregates the encrypted models, and a new federation round begins.}
    \label{fig:HETrainPipeline}
\end{figure*}

\paragraph{Secure FL using FHE.} Figure~\ref{fig:HETrainPipeline} presents the secure federated training pipeline of our MetisFL system. We use homomorphic encryption to communicate the (encrypted) local and global models between the Federation Controller and the learners and compute the new global model by aggregating learners' local models in an encrypted space. Training starts with an initial configuration phase, where the Federation Driver generates the homomorphic encryption key pair (private and public key) and the original neural model state. The Federation Controller only receives the model definition and the public key from the driver since it only needs to perform the private weighted aggregation of local models. On the contrary, the learners need private and public keys during training. The private key is used to decrypt the encrypted global model received by the controller to perform their local training (or model evaluation) over their local private dataset, and the public key is used to encrypt the locally trained model before being shared with the controller. 

We used a similar training pipeline in Stripelis et al.~\cite{stripelis2021secure}. However, in our previous work, we encrypted the entire model into a single ciphertext, which created scalability issues for large models. To mitigate this, in MetisFL we encrypt the model on a matrix-by-matrix basis, allowing for a collection of ciphertexts to be communicated between learners and the controller instead of just a single ciphertext. Thereafter, the controller performs the private weighted aggregation over the collection of ciphertexts from all learners. Additionally, we divide the model parameters into batches processed in parallel, leading to a much faster encrypted computation. 

\begin{figure}
    \centering
    \includegraphics[width=0.5\linewidth]{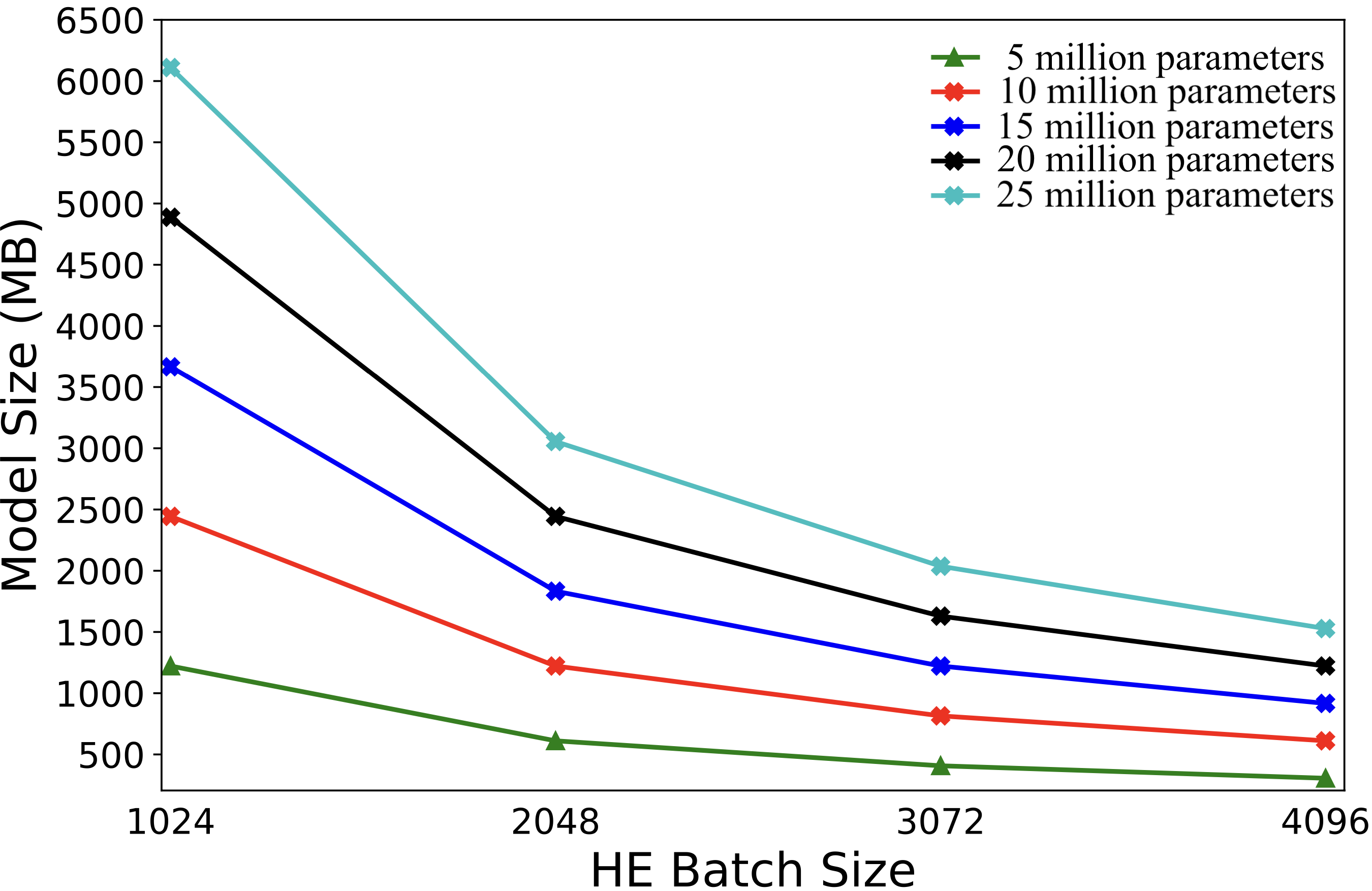}
    \caption{Relation between the encrypted model size and the batch size of the CKKS scheme. The encrypted model size is reduced as we increase the batch size of the FHE scheme (scaling factor bits = 52, security level = 128).}%
    \label{fig:model_size_vs_he_batch_size}
\end{figure}

Figure~\ref{fig:model_size_vs_he_batch_size} demonstrates the effect of batching multiple model parameters on the size of the encrypted model. Batching multiple parameters into a single ciphertext helps us reduce the overall model size and allows us to leverage SIMD (Single Instruction Multiple Data) for faster processing of encrypted models. The CKKS parameters are multiplicative depth of 1, 52 scale factor bits, batch size of 4096, and security level of 128 bits. 

\paragraph{Membership Inference Attacks.} To conduct the membership inference attacks we use the same features and architecture as in Gupta et al.~\cite{gupta2021membership} for training models to predict membership. We train the attack model for each learner by creating a training set from their training samples and  samples not seen during the model training. Finally, we compute how accurately we can predict the training samples of all the other learners vs.\ samples not used for training. We create a balanced set of samples used for training and unseeen samples  (i.e., the test samples) from each learner to compute this accuracy. We report the average of these 56 accuracy values as the vulnerability score --- each learner trains attack models (i.e., 8 attackers) and predicts train vs.\ unseen on samples from other 7 learners.

\textit{Gaussian Noise.} Differential privacy is a formal framework to reason about privacy. A differential private training mechanism ensures that the outcomes (i.e., final model weights) do not change much between two training sets that differ by one example~\cite{dwork2014algorithmic}. For training brain age models with differential privacy guarantees, we use the DP-SGD algorithm~\cite{abadi2016deep}. Briefly, the principal modifications to SGD to limit the influence of a single sample are to clip the gradients from each sample not to exceed a maximum norm and to add spherical Gaussian noise. We update each learner during federated training with these private gradients. During initial experiments, we found that achieving non-vacuous differential privacy guarantees requires adding significant Gaussian noise to the gradients, which annihilates learning performance. However, we observed that practical privacy attacks, such as membership inference attacks, can be thwarted by adding Gaussian noise of much smaller magnitudes~\cite{jayaraman2019evaluating}. Therefore, we evaluate training with gradients with a small additive Gaussian noise. 

\textit{Learning with Non-unique Gradients.} To learn good machine learning models, we would like to extract patterns while ignoring information about specific samples. Training models using gradient descent can leak an individual’s information during training because there is no restriction on what information a sample may contribute. Thus, the model may preserve information unique to each individual, leaking privacy. Differential privacy adds the same noise to all gradients to limit the information or influence of a single sample on the neural network, but that may also destroy useful information in the attempt to reduce memorization. %
We investigate removing unique information from each sample's gradient and training with only non-unique parts. We compute the gradient of the loss ($L$) w.r.t. parameters ($\theta$) for each sample ($x_i,y_i$) in a batch ($B$), i.e., $g_i=\nabla_\theta L(f(x_i;\theta), y_i)\ \forall i\in\{1\ldots B\}$. To compute the non-unique part, we project each gradient vector on the subspace spanned by the rest of the gradient vectors ($g_i^{span}$). We consider the residual part as the unique information about each sample (i.e., $g_i^{unique} = g_i-g_i^{span}$. Ideally, we would like to train with only the non-unique part. However, we observe that it may harm the performance too much, and therefore we downweigh the effect of the unique part and use $\hat g_i=g_i^{span}+\alpha g_i^{unique}, \alpha<1$ to update the model at local learners; $\alpha$ is a hyperparameter that we tune to trade-off privacy and performance.

We hypothesize that the small additive noise is enough to reduce the mutual information between data and neural network outputs/activations, which limits the success of membership inference attacks~\cite{jha2020extension}. 
In IID environments, non-unique gradients perform similarly to adding Gaussian noise. However, they are significantly faster to train. The Gaussian noise models required training for 40 rounds, whereas the non-unique gradients required only 25 rounds. 
In Non-IID environments, training with Gaussian noise provides a better trade-off than non-unique gradients. This may be due to learners overfitting to private datasets earlier in training, thus deviating from the community model. 
In summary, both small-magnitude Gaussian noise added to Gradients, and non-unique gradients are effective at preventing membership attacks.
\section*{Acknowledgements}
This research was supported in part by the Defense Advanced Research Projects Agency (DARPA) under contract HR00112090104, and in part by the National Institutes of Health (NIH) under grants U01AG068057 and RF1AG051710. The views and conclusions contained herein are those of the authors and should not be interpreted as necessarily representing the official policies or endorsements, either expressed or implied, of DARPA, NIH, or the U.S. Government.

\section*{Author contributions statement}
D.S. co-led the effort, developed the system architecture, experiments, and analyzed findings. 
U.G. developed the privacy components and analyzed the findings.
N.D. defined the deep learning models used for the BrainAGE and AD experiments.
H.S. and T.G. helped with the homomorphic encryption development. 
C.A. and A.A. helped with the Metis federated learning system design and development.
G.V.S. advised on the data privacy analysis.
S.R. and M.N. advised on system security development.
P.M.T. advised on the neuroimaging data analysis and experiments.
J.L.A. led the effort, co-designed the federated learning architecture and training algorithms, and advised on experiment design and data analysis findings. 
Every author has approved the submitted version and has agreed both to be personally accountable for the author’s own contributions and to ensure that questions related to the accuracy or integrity of any part of the work, even ones in which the author was not personally involved, are appropriately investigated, resolved, and the resolution documented in the literature.

\section*{Data availability statement}
This is a study of previously collected, anonymized, de-identified data, available in a public repository. For the UK Biobank data access was approved by UK Biobank under Application Number 11559. All other datasets analyzed during the current study are available in the ADNI [\url{https://adni.loni.usc.edu/}], OASIS-3 [\url{https://www.oasis-brains.org/}], and AIBL [\url{https://aibl.csiro.au/}] repositories. Further information is available in the following published articles for ADNI~\cite{veitch2019understanding} and OASIS-3~\cite{LaMontagne2019.12.13.19014902}.

\section*{Competing interests}
The authors declare no competing interests.

\section*{Additional information}
\noindent\textbf{Correspondence} and requests for materials should be addressed to Dimitris Stripelis. %

\bibliography{main}

\begin{thebibliography}{10}
\urlstyle{rm}
\expandafter\ifx\csname url\endcsname\relax
  \def\url#1{\texttt{#1}}\fi
\expandafter\ifx\csname urlprefix\endcsname\relax\def\urlprefix{URL }\fi
\expandafter\ifx\csname doiprefix\endcsname\relax\def\doiprefix{DOI: }\fi
\providecommand{\bibinfo}[2]{#2}
\providecommand{\eprint}[2][]{\url{#2}}

\bibitem{wainberg2018deep}
\bibinfo{author}{Wainberg, M.}, \bibinfo{author}{Merico, D.},
  \bibinfo{author}{Delong, A.} \& \bibinfo{author}{Frey, B.~J.}
\newblock \bibinfo{journal}{\bibinfo{title}{Deep learning in biomedicine}}.
\newblock {\emph{\JournalTitle{Nature biotechnology}}}
  \textbf{\bibinfo{volume}{36}}, \bibinfo{pages}{829--838}
  (\bibinfo{year}{2018}).

\bibitem{suzuki2017overview}
\bibinfo{author}{Suzuki, K.}
\newblock \bibinfo{journal}{\bibinfo{title}{Overview of deep learning in
  medical imaging}}.
\newblock {\emph{\JournalTitle{Radiological physics and technology}}}
  \textbf{\bibinfo{volume}{10}}, \bibinfo{pages}{257--273}
  (\bibinfo{year}{2017}).

\bibitem{zhu2018image}
\bibinfo{author}{Zhu, B.}, \bibinfo{author}{Liu, J.~Z.},
  \bibinfo{author}{Cauley, S.~F.}, \bibinfo{author}{Rosen, B.~R.} \&
  \bibinfo{author}{Rosen, M.~S.}
\newblock \bibinfo{journal}{\bibinfo{title}{Image reconstruction by
  domain-transform manifold learning}}.
\newblock {\emph{\JournalTitle{Nature}}} \textbf{\bibinfo{volume}{555}},
  \bibinfo{pages}{487--492} (\bibinfo{year}{2018}).

\bibitem{dalca2019unsupervised}
\bibinfo{author}{Dalca, A.~V.} \emph{et~al.}
\newblock \bibinfo{title}{Unsupervised deep learning for bayesian brain mri
  segmentation}.
\newblock In \emph{\bibinfo{booktitle}{International Conference on Medical
  Image Computing and Computer-Assisted Intervention}},
  \bibinfo{pages}{356--365} (\bibinfo{organization}{Springer},
  \bibinfo{year}{2019}).

\bibitem{cho2021deep}
\bibinfo{author}{Cho, J.} \emph{et~al.}
\newblock \bibinfo{journal}{\bibinfo{title}{Deep learning-based computer-aided
  detection system for automated treatment response assessment of brain
  metastases on 3d mri}}.
\newblock {\emph{\JournalTitle{Frontiers in Oncology}}} \bibinfo{pages}{4314}
  (\bibinfo{year}{2021}).

\bibitem{kofler2020brats}
\bibinfo{author}{Kofler, F.} \emph{et~al.}
\newblock \bibinfo{journal}{\bibinfo{title}{Brats toolkit: translating brats
  brain tumor segmentation algorithms into clinical and scientific practice}}.
\newblock {\emph{\JournalTitle{Frontiers in neuroscience}}}
  \bibinfo{pages}{125} (\bibinfo{year}{2020}).

\bibitem{aksman2021pysustain}
\bibinfo{author}{Aksman, L.~M.} \emph{et~al.}
\newblock \bibinfo{journal}{\bibinfo{title}{pysustain: a python implementation
  of the subtype and stage inference algorithm}}.
\newblock {\emph{\JournalTitle{SoftwareX}}} \textbf{\bibinfo{volume}{16}},
  \bibinfo{pages}{100811} (\bibinfo{year}{2021}).

\bibitem{young2021ordinal}
\bibinfo{author}{Young, A.~L.} \emph{et~al.}
\newblock \bibinfo{journal}{\bibinfo{title}{Ordinal sustain: Subtype and stage
  inference for clinical scores, visual ratings, and other ordinal data}}.
\newblock {\emph{\JournalTitle{Frontiers in artificial intelligence}}}
  \textbf{\bibinfo{volume}{4}} (\bibinfo{year}{2021}).

\bibitem{ezzati2021predictive}
\bibinfo{author}{Ezzati, A.} \emph{et~al.}
\newblock \bibinfo{journal}{\bibinfo{title}{Predictive value of atn biomarker
  profiles in estimating disease progression in alzheimer's disease dementia}}.
\newblock {\emph{\JournalTitle{Alzheimer's \& Dementia}}}
  \textbf{\bibinfo{volume}{17}}, \bibinfo{pages}{1855--1867}
  (\bibinfo{year}{2021}).

\bibitem{lu2021practical}
\bibinfo{author}{Lu, B.} \emph{et~al.}
\newblock \bibinfo{journal}{\bibinfo{title}{A practical alzheimer disease
  classifier via brain imaging-based deep learning on 85,721 samples: A
  multicentre, retrospective cohort study}}.
\newblock {\emph{\JournalTitle{BioRxiv}}} \bibinfo{pages}{2020--08}
  (\bibinfo{year}{2021}).

\bibitem{thompson2020enigma}
\bibinfo{author}{Thompson, P.~M.} \emph{et~al.}
\newblock \bibinfo{journal}{\bibinfo{title}{Enigma and global neuroscience: A
  decade of large-scale studies of the brain in health and disease across more
  than 40 countries}}.
\newblock {\emph{\JournalTitle{Translational psychiatry}}}
  \textbf{\bibinfo{volume}{10}}, \bibinfo{pages}{1--28} (\bibinfo{year}{2020}).

\bibitem{bischoff2007technique}
\bibinfo{author}{Bischoff-Grethe, A.} \emph{et~al.}
\newblock \bibinfo{journal}{\bibinfo{title}{{A technique for the
  deidentification of structural brain MR images}}}.
\newblock {\emph{\JournalTitle{Human brain mapping}}}
  \textbf{\bibinfo{volume}{28}}, \bibinfo{pages}{892--903}
  (\bibinfo{year}{2007}).

\bibitem{schimke2011quickshear}
\bibinfo{author}{Schimke, N.} \& \bibinfo{author}{Hale, J.}
\newblock \bibinfo{title}{{Quickshear Defacing for Neuroimages}}.
\newblock In \emph{\bibinfo{booktitle}{Proceedings of the 2nd USENIX conference
  on Health security and privacy}}, \bibinfo{pages}{11--11}
  (\bibinfo{organization}{USENIX Association}, \bibinfo{year}{2011}).

\bibitem{milchenko2013obscuring}
\bibinfo{author}{Milchenko, M.} \& \bibinfo{author}{Marcus, D.}
\newblock \bibinfo{journal}{\bibinfo{title}{{Obscuring Surface Anatomy in
  Volumetric Imaging Data }}}.
\newblock {\emph{\JournalTitle{Neuroinformatics}}}
  \textbf{\bibinfo{volume}{11}}, \bibinfo{pages}{65--75}
  (\bibinfo{year}{2013}).

\bibitem{mcmahan2017communication}
\bibinfo{author}{McMahan, B.}, \bibinfo{author}{Moore, E.},
  \bibinfo{author}{Ramage, D.}, \bibinfo{author}{Hampson, S.} \&
  \bibinfo{author}{y~Arcas, B.~A.}
\newblock \bibinfo{title}{Communication-efficient learning of deep networks
  from decentralized data}.
\newblock In \emph{\bibinfo{booktitle}{Artificial Intelligence and
  Statistics}}, \bibinfo{pages}{1273--1282} (\bibinfo{organization}{PMLR},
  \bibinfo{year}{2017}).

\bibitem{yang2019federated}
\bibinfo{author}{Yang, Q.}, \bibinfo{author}{Liu, Y.}, \bibinfo{author}{Chen,
  T.} \& \bibinfo{author}{Tong, Y.}
\newblock \bibinfo{journal}{\bibinfo{title}{Federated machine learning: Concept
  and applications}}.
\newblock {\emph{\JournalTitle{ACM Transactions on Intelligent Systems and
  Technology (TIST)}}} \textbf{\bibinfo{volume}{10}}, \bibinfo{pages}{1--19}
  (\bibinfo{year}{2019}).

\bibitem{li2020federated}
\bibinfo{author}{Li, T.}, \bibinfo{author}{Sahu, A.~K.},
  \bibinfo{author}{Talwalkar, A.} \& \bibinfo{author}{Smith, V.}
\newblock \bibinfo{journal}{\bibinfo{title}{Federated learning: Challenges,
  methods, and future directions}}.
\newblock {\emph{\JournalTitle{IEEE Signal Processing Magazine}}}
  \textbf{\bibinfo{volume}{37}}, \bibinfo{pages}{50--60}
  (\bibinfo{year}{2020}).

\bibitem{lee2018privacy}
\bibinfo{author}{Lee, J.} \emph{et~al.}
\newblock \bibinfo{journal}{\bibinfo{title}{Privacy-preserving patient
  similarity learning in a federated environment: development and analysis}}.
\newblock {\emph{\JournalTitle{JMIR medical informatics}}}
  \textbf{\bibinfo{volume}{6}}, \bibinfo{pages}{e7744} (\bibinfo{year}{2018}).

\bibitem{sheller2018multi}
\bibinfo{author}{Sheller, M.~J.}, \bibinfo{author}{Reina, G.~A.},
  \bibinfo{author}{Edwards, B.}, \bibinfo{author}{Martin, J.} \&
  \bibinfo{author}{Bakas, S.}
\newblock \bibinfo{title}{Multi-institutional deep learning modeling without
  sharing patient data: A feasibility study on brain tumor segmentation}.
\newblock In \emph{\bibinfo{booktitle}{International MICCAI Brainlesion
  Workshop}}, \bibinfo{pages}{92--104} (\bibinfo{organization}{Springer},
  \bibinfo{year}{2018}).

\bibitem{silva2019federated}
\bibinfo{author}{Silva, S.} \emph{et~al.}
\newblock \bibinfo{title}{Federated learning in distributed medical databases:
  Meta-analysis of large-scale subcortical brain data}.
\newblock In \emph{\bibinfo{booktitle}{2019 IEEE 16th international symposium
  on biomedical imaging (ISBI 2019)}}, \bibinfo{pages}{270--274}
  (\bibinfo{organization}{IEEE}, \bibinfo{year}{2019}).

\bibitem{rieke2020future}
\bibinfo{author}{Rieke, N.} \emph{et~al.}
\newblock \bibinfo{journal}{\bibinfo{title}{The future of digital health with
  federated learning}}.
\newblock {\emph{\JournalTitle{NPJ digital medicine}}}
  \textbf{\bibinfo{volume}{3}}, \bibinfo{pages}{1--7} (\bibinfo{year}{2020}).

\bibitem{silva2020fed}
\bibinfo{author}{Silva, S.}, \bibinfo{author}{Altmann, A.},
  \bibinfo{author}{Gutman, B.} \& \bibinfo{author}{Lorenzi, M.}
\newblock \bibinfo{title}{Fed-biomed: A general open-source frontend framework
  for federated learning in healthcare}.
\newblock In \emph{\bibinfo{booktitle}{Domain Adaptation and Representation
  Transfer, and Distributed and Collaborative Learning}},
  \bibinfo{pages}{201--210} (\bibinfo{publisher}{Springer},
  \bibinfo{year}{2020}).

\bibitem{stripelis2021secure}
\bibinfo{author}{Stripelis, D.} \emph{et~al.}
\newblock \bibinfo{title}{Secure neuroimaging analysis using federated learning
  with homomorphic encryption}.
\newblock In \emph{\bibinfo{booktitle}{17th International Symposium on Medical
  Information Processing and Analysis}}, vol. \bibinfo{volume}{12088},
  \bibinfo{pages}{351--359} (\bibinfo{organization}{SPIE},
  \bibinfo{year}{2021}).

\bibitem{ckks_paper}
\bibinfo{author}{Cheon, J.~H.}, \bibinfo{author}{Kim, A.},
  \bibinfo{author}{Kim, M.} \& \bibinfo{author}{Song, Y.}
\newblock \bibinfo{title}{Homomorphic encryption for arithmetic of approximate
  numbers}.
\newblock In \bibinfo{editor}{Takagi, T.} \& \bibinfo{editor}{Peyrin, T.}
  (eds.) \emph{\bibinfo{booktitle}{Advances in Cryptology -- ASIACRYPT 2017}},
  \bibinfo{pages}{409--437} (\bibinfo{year}{2017}).

\bibitem{geiping2020inverting}
\bibinfo{author}{Geiping, J.}, \bibinfo{author}{Bauermeister, H.},
  \bibinfo{author}{Dr{\"o}ge, H.} \& \bibinfo{author}{Moeller, M.}
\newblock \bibinfo{journal}{\bibinfo{title}{Inverting gradients-how easy is it
  to break privacy in federated learning?}}
\newblock {\emph{\JournalTitle{Advances in Neural Information Processing
  Systems}}} \textbf{\bibinfo{volume}{33}}, \bibinfo{pages}{16937--16947}
  (\bibinfo{year}{2020}).

\bibitem{zhu2019deep}
\bibinfo{author}{Zhu, L.}, \bibinfo{author}{Liu, Z.} \& \bibinfo{author}{Han,
  S.}
\newblock \bibinfo{journal}{\bibinfo{title}{Deep leakage from gradients}}.
\newblock {\emph{\JournalTitle{Advances in Neural Information Processing
  Systems}}} \textbf{\bibinfo{volume}{32}} (\bibinfo{year}{2019}).

\bibitem{shokri2017}
\bibinfo{author}{Shokri, R.}, \bibinfo{author}{Stronati, M.},
  \bibinfo{author}{Song, C.} \& \bibinfo{author}{Shmatikov, V.}
\newblock \bibinfo{title}{{Membership Inference Attacks Against Machine
  Learning Models}}.
\newblock In \emph{\bibinfo{booktitle}{2017 IEEE Symposium on Security and
  Privacy (SP)}}, \bibinfo{pages}{3--18} (\bibinfo{year}{2017}).

\bibitem{nasr2019}
\bibinfo{author}{{Nasr}, M.}, \bibinfo{author}{{Shokri}, R.} \&
  \bibinfo{author}{{Houmansadr}, A.}
\newblock \bibinfo{title}{{Comprehensive Privacy Analysis of Deep Learning:
  Passive and Active White-box Inference Attacks against Centralized and
  Federated Learning}}.
\newblock In \emph{\bibinfo{booktitle}{2019 IEEE Symposium on Security and
  Privacy (SP)}}, \bibinfo{pages}{739--753} (\bibinfo{year}{2019}).

\bibitem{miller2016multimodal}
\bibinfo{author}{Miller, K.~L.} \emph{et~al.}
\newblock \bibinfo{journal}{\bibinfo{title}{Multimodal population brain imaging
  in the uk biobank prospective epidemiological study}}.
\newblock {\emph{\JournalTitle{Nature neuroscience}}}
  \textbf{\bibinfo{volume}{19}}, \bibinfo{pages}{1523--1536}
  (\bibinfo{year}{2016}).

\bibitem{mueller2005alzheimer}
\bibinfo{author}{Mueller, S.~G.} \emph{et~al.}
\newblock \bibinfo{journal}{\bibinfo{title}{The alzheimer's disease
  neuroimaging initiative}}.
\newblock {\emph{\JournalTitle{Neuroimaging Clinics}}}
  \textbf{\bibinfo{volume}{15}}, \bibinfo{pages}{869--877}
  (\bibinfo{year}{2005}).

\bibitem{LaMontagne2019.12.13.19014902}
\bibinfo{author}{LaMontagne, P.~J.} \emph{et~al.}
\newblock \bibinfo{journal}{\bibinfo{title}{Oasis-3: Longitudinal neuroimaging,
  clinical, and cognitive dataset for normal aging and alzheimer disease}}.
\newblock {\emph{\JournalTitle{medRxiv}}}
  \doiprefix\url{10.1101/2019.12.13.19014902} (\bibinfo{year}{2019}).
\newblock
  \eprint{https://www.medrxiv.org/content/early/2019/12/15/2019.12.13.19014902.full.pdf}.

\bibitem{fowler2021fifteen}
\bibinfo{author}{Fowler, C.} \emph{et~al.}
\newblock \bibinfo{journal}{\bibinfo{title}{Fifteen years of the australian
  imaging, biomarkers and lifestyle (aibl) study: progress and observations
  from 2,359 older adults spanning the spectrum from cognitive normality to
  alzheimer’s disease}}.
\newblock {\emph{\JournalTitle{Journal of Alzheimer's disease reports}}}
  \bibinfo{pages}{1--26} (\bibinfo{year}{2021}).

\bibitem{hatamizadeh2022gradient}
\bibinfo{author}{Hatamizadeh, A.} \emph{et~al.}
\newblock \bibinfo{journal}{\bibinfo{title}{Do gradient inversion attacks make
  federated learning unsafe?}}
\newblock {\emph{\JournalTitle{arXiv preprint arXiv:2202.06924}}}
  (\bibinfo{year}{2022}).

\bibitem{gupta2021membership}
\bibinfo{author}{Gupta, U.} \emph{et~al.}
\newblock \bibinfo{title}{Membership inference attacks on deep regression
  models for neuroimaging}.
\newblock In \emph{\bibinfo{booktitle}{Medical Imaging with Deep Learning}},
  \bibinfo{pages}{228--251} (\bibinfo{organization}{PMLR},
  \bibinfo{year}{2021}).

\bibitem{yeom2018privacy}
\bibinfo{author}{Yeom, S.}, \bibinfo{author}{Giacomelli, I.},
  \bibinfo{author}{Fredrikson, M.} \& \bibinfo{author}{Jha, S.}
\newblock \bibinfo{title}{Privacy risk in machine learning: Analyzing the
  connection to overfitting}.
\newblock In \emph{\bibinfo{booktitle}{2018 IEEE 31st computer security
  foundations symposium (CSF)}}, \bibinfo{pages}{268--282}
  (\bibinfo{organization}{IEEE}, \bibinfo{year}{2018}).

\bibitem{truex2018towards}
\bibinfo{author}{Truex, S.}, \bibinfo{author}{Liu, L.},
  \bibinfo{author}{Gursoy, M.~E.}, \bibinfo{author}{Yu, L.} \&
  \bibinfo{author}{Wei, W.}
\newblock \bibinfo{journal}{\bibinfo{title}{{Towards Demystifying Membership
  Inference Attacks}}}.
\newblock {\emph{\JournalTitle{arXiv preprint arXiv:1807.09173}}}
  (\bibinfo{year}{2018}).

\bibitem{salem2019ml}
\bibinfo{author}{Salem, A.}, \bibinfo{author}{Zhang, Y.},
  \bibinfo{author}{Humbert, M.}, \bibinfo{author}{Fritz, M.} \&
  \bibinfo{author}{Backes, M.}
\newblock \bibinfo{title}{{ML-Leaks: Model and Data Independent Membership
  Inference Attacks and Defenses on Machine Learning Models}}.
\newblock In \emph{\bibinfo{booktitle}{Network and Distributed Systems Security
  Symposium 2019}} (\bibinfo{organization}{Internet Society},
  \bibinfo{year}{2019}).

\bibitem{jha2020extension}
\bibinfo{author}{Jha, S.~K.} \emph{et~al.}
\newblock \bibinfo{journal}{\bibinfo{title}{{An Extension of Fano's Inequality
  for Characterizing Model Susceptibility to Membership Inference Attacks}}}.
\newblock {\emph{\JournalTitle{arXiv preprint arXiv:2009.08097}}}
  (\bibinfo{year}{2020}).

\bibitem{abadi2016deep}
\bibinfo{author}{Abadi, M.} \emph{et~al.}
\newblock \bibinfo{title}{{Deep Learning with Differential Privacy}}.
\newblock In \emph{\bibinfo{booktitle}{Proceedings of the 2016 ACM SIGSAC
  conference on computer and communications security}},
  \bibinfo{pages}{308--318} (\bibinfo{year}{2016}).

\bibitem{patterson2018world}
\bibinfo{author}{Patterson, C.}
\newblock \bibinfo{title}{World alzheimer report 2018} (\bibinfo{year}{2018}).

\bibitem{10.1117/12.2606297}
\bibinfo{author}{Dhinagar, N.~J.} \emph{et~al.}
\newblock \bibinfo{title}{{3D convolutional neural networks for classification
  of Alzheimer’s and Parkinson’s disease with T1-weighted brain MRI}}.
\newblock In \bibinfo{editor}{Rittner, L.}, \bibinfo{editor}{M.D., E. R.~C.},
  \bibinfo{editor}{Lepore, N.}, \bibinfo{editor}{Brieva, J.} \&
  \bibinfo{editor}{Linguraru, M.~G.} (eds.) \emph{\bibinfo{booktitle}{17th
  International Symposium on Medical Information Processing and Analysis}},
  vol. \bibinfo{volume}{12088}, \bibinfo{pages}{277 -- 286},
  \doiprefix\url{10.1117/12.2606297}. \bibinfo{organization}{International
  Society for Optics and Photonics} (\bibinfo{publisher}{SPIE},
  \bibinfo{year}{2021}).

\bibitem{abdulazeem2021cnn}
\bibinfo{author}{AbdulAzeem, Y.}, \bibinfo{author}{Bahgat, W.~M.} \&
  \bibinfo{author}{Badawy, M.}
\newblock \bibinfo{journal}{\bibinfo{title}{A cnn based framework for
  classification of alzheimer’s disease}}.
\newblock {\emph{\JournalTitle{Neural Computing and Applications}}}
  \textbf{\bibinfo{volume}{33}}, \bibinfo{pages}{10415--10428}
  (\bibinfo{year}{2021}).

\bibitem{fu2021automated}
\bibinfo{author}{Fu’adah, Y.} \emph{et~al.}
\newblock \bibinfo{title}{Automated classification of alzheimer’s disease
  based on mri image processing using convolutional neural network (cnn) with
  alexnet architecture}.
\newblock In \emph{\bibinfo{booktitle}{Journal of Physics: Conference Series}},
  vol. \bibinfo{volume}{1844}, \bibinfo{pages}{012020}
  (\bibinfo{organization}{IOP Publishing}, \bibinfo{year}{2021}).

\bibitem{franke2019ten}
\bibinfo{author}{Franke, K.} \& \bibinfo{author}{Gaser, C.}
\newblock \bibinfo{journal}{\bibinfo{title}{Ten years of brainage as a
  neuroimaging biomarker of brain aging: what insights have we gained?}}
\newblock {\emph{\JournalTitle{Frontiers in neurology}}} \bibinfo{pages}{789}
  (\bibinfo{year}{2019}).

\bibitem{wood2022accurate}
\bibinfo{author}{Wood, D.~A.} \emph{et~al.}
\newblock \bibinfo{journal}{\bibinfo{title}{Accurate brain-age models for
  routine clinical mri examinations}}.
\newblock {\emph{\JournalTitle{NeuroImage}}} \bibinfo{pages}{118871}
  (\bibinfo{year}{2022}).

\bibitem{cole2015prediction}
\bibinfo{author}{Cole, J.~H.}, \bibinfo{author}{Leech, R.},
  \bibinfo{author}{Sharp, D.~J.} \& \bibinfo{author}{Initiative, A. D.~N.}
\newblock \bibinfo{journal}{\bibinfo{title}{Prediction of brain age suggests
  accelerated atrophy after traumatic brain injury}}.
\newblock {\emph{\JournalTitle{Annals of neurology}}}
  \textbf{\bibinfo{volume}{77}}, \bibinfo{pages}{571--581}
  (\bibinfo{year}{2015}).

\bibitem{koutsouleris2014accelerated}
\bibinfo{author}{Koutsouleris, N.} \emph{et~al.}
\newblock \bibinfo{journal}{\bibinfo{title}{Accelerated brain aging in
  schizophrenia and beyond: a neuroanatomical marker of psychiatric
  disorders}}.
\newblock {\emph{\JournalTitle{Schizophrenia bulletin}}}
  \textbf{\bibinfo{volume}{40}}, \bibinfo{pages}{1140--1153}
  (\bibinfo{year}{2014}).

\bibitem{kuchinad2007accelerated}
\bibinfo{author}{Kuchinad, A.} \emph{et~al.}
\newblock \bibinfo{journal}{\bibinfo{title}{Accelerated brain gray matter loss
  in fibromyalgia patients: premature aging of the brain?}}
\newblock {\emph{\JournalTitle{Journal of Neuroscience}}}
  \textbf{\bibinfo{volume}{27}}, \bibinfo{pages}{4004--4007}
  (\bibinfo{year}{2007}).

\bibitem{lam2020accurate}
\bibinfo{author}{Lam, P.~K.} \emph{et~al.}
\newblock \bibinfo{title}{Accurate brain age prediction using recurrent
  slice-based networks}.
\newblock In \emph{\bibinfo{booktitle}{16th International Symposium on Medical
  Information Processing and Analysis}}, vol. \bibinfo{volume}{11583},
  \bibinfo{pages}{1158303} (\bibinfo{organization}{International Society for
  Optics and Photonics}, \bibinfo{year}{2020}).

\bibitem{jonsson2019brain}
\bibinfo{author}{J{\'o}nsson, B.~A.} \emph{et~al.}
\newblock \bibinfo{journal}{\bibinfo{title}{Brain age prediction using deep
  learning uncovers associated sequence variants}}.
\newblock {\emph{\JournalTitle{Nature Communications}}}
  \textbf{\bibinfo{volume}{10}}, \bibinfo{pages}{1--10} (\bibinfo{year}{2019}).

\bibitem{dinsdale2021learning}
\bibinfo{author}{Dinsdale, N.~K.} \emph{et~al.}
\newblock \bibinfo{journal}{\bibinfo{title}{Learning patterns of the ageing
  brain in mri using deep convolutional networks}}.
\newblock {\emph{\JournalTitle{NeuroImage}}} \textbf{\bibinfo{volume}{224}},
  \bibinfo{pages}{117401} (\bibinfo{year}{2021}).

\bibitem{cole2017predicting}
\bibinfo{author}{Cole, J.~H.} \emph{et~al.}
\newblock \bibinfo{journal}{\bibinfo{title}{Predicting brain age with deep
  learning from raw imaging data results in a reliable and heritable
  biomarker}}.
\newblock {\emph{\JournalTitle{NeuroImage}}} \textbf{\bibinfo{volume}{163}},
  \bibinfo{pages}{115--124} (\bibinfo{year}{2017}).

\bibitem{peng2021accurate}
\bibinfo{author}{Peng, H.}, \bibinfo{author}{Gong, W.},
  \bibinfo{author}{Beckmann, C.~F.}, \bibinfo{author}{Vedaldi, A.} \&
  \bibinfo{author}{Smith, S.~M.}
\newblock \bibinfo{journal}{\bibinfo{title}{Accurate brain age prediction with
  lightweight deep neural networks}}.
\newblock {\emph{\JournalTitle{Medical image analysis}}}
  \textbf{\bibinfo{volume}{68}}, \bibinfo{pages}{101871}
  (\bibinfo{year}{2021}).

\bibitem{gupta2021improved}
\bibinfo{author}{Gupta, U.}, \bibinfo{author}{Lam, P.~K.},
  \bibinfo{author}{Ver~Steeg, G.} \& \bibinfo{author}{Thompson, P.~M.}
\newblock \bibinfo{title}{Improved brain age estimation with slice-based set
  networks}.
\newblock In \emph{\bibinfo{booktitle}{2021 IEEE 18th International Symposium
  on Biomedical Imaging (ISBI)}}, \bibinfo{pages}{840--844}
  (\bibinfo{organization}{IEEE}, \bibinfo{year}{2021}).

\bibitem{stripelis2021scaling}
\bibinfo{author}{Stripelis, D.}, \bibinfo{author}{Ambite, J.~L.},
  \bibinfo{author}{Lam, P.} \& \bibinfo{author}{Thompson, P.}
\newblock \bibinfo{title}{Scaling neuroscience research using federated
  learning}.
\newblock In \emph{\bibinfo{booktitle}{2021 IEEE 18th International Symposium
  on Biomedical Imaging (ISBI)}}, \bibinfo{pages}{1191--1195}
  (\bibinfo{organization}{IEEE}, \bibinfo{year}{2021}).

\bibitem{stripelis2022semi}
\bibinfo{author}{Stripelis, D.}, \bibinfo{author}{Ambite, J.~L.} \&
  \bibinfo{author}{Paul, T.}
\newblock \bibinfo{journal}{\bibinfo{title}{Semi-synchronous federated learning
  for energy-efficient training and accelerated convergence in cross-silo
  settings}}.
\newblock {\emph{\JournalTitle{ACM Transactions on Intelligent Systems and
  Technology, Special Issue on Federated Learning: Algorithms, Systems, and
  Applications. Forthcoming}}} \doiprefix\url{10.1145/3524885}
  (\bibinfo{year}{2022}).

\bibitem{wei2020federated}
\bibinfo{author}{Wei, K.} \emph{et~al.}
\newblock \bibinfo{journal}{\bibinfo{title}{Federated learning with
  differential privacy: Algorithms and performance analysis}}.
\newblock {\emph{\JournalTitle{IEEE Transactions on Information Forensics and
  Security}}} \textbf{\bibinfo{volume}{15}}, \bibinfo{pages}{3454--3469}
  (\bibinfo{year}{2020}).

\bibitem{noble2021differentially}
\bibinfo{author}{Noble, M.}, \bibinfo{author}{Bellet, A.} \&
  \bibinfo{author}{Dieuleveut, A.}
\newblock \bibinfo{journal}{\bibinfo{title}{Differentially private federated
  learning on heterogeneous data}}.
\newblock {\emph{\JournalTitle{arXiv preprint arXiv:2111.09278}}}
  (\bibinfo{year}{2021}).

\bibitem{zhao2020local}
\bibinfo{author}{Zhao, Y.} \emph{et~al.}
\newblock \bibinfo{journal}{\bibinfo{title}{Local differential privacy-based
  federated learning for internet of things}}.
\newblock {\emph{\JournalTitle{IEEE Internet of Things Journal}}}
  \textbf{\bibinfo{volume}{8}}, \bibinfo{pages}{8836--8853}
  (\bibinfo{year}{2020}).

\bibitem{HomomorphicEncryptionSecurityStandard}
\bibinfo{author}{Albrecht, M.} \emph{et~al.}
\newblock \bibinfo{title}{Homomorphic encryption security standard}.
\newblock \bibinfo{type}{Tech. Rep.},
  \bibinfo{institution}{HomomorphicEncryption.org}, \bibinfo{address}{Toronto,
  Canada} (\bibinfo{year}{2018}).

\bibitem{Sako2011}
\bibinfo{author}{Sako, K.}
\newblock \emph{\bibinfo{title}{Public Key Cryptography}},
  \bibinfo{pages}{996--997} (\bibinfo{publisher}{Springer US},
  \bibinfo{address}{Boston, MA}, \bibinfo{year}{2011}).

\bibitem{DBLP:journals/corr/abs-1812-03224}
\bibinfo{author}{Truex, S.} \emph{et~al.}
\newblock \bibinfo{journal}{\bibinfo{title}{A hybrid approach to
  privacy-preserving federated learning}}.
\newblock {\emph{\JournalTitle{CoRR}}}
  \textbf{\bibinfo{volume}{abs/1812.03224}} (\bibinfo{year}{2018}).
\newblock \eprint{1812.03224}.

\bibitem{batchcrypt_paper}
\bibinfo{author}{Zhang, C.} \emph{et~al.}
\newblock \bibinfo{title}{Batchcrypt: Efficient homomorphic encryption for
  cross-silo federated learning}.
\newblock In \emph{\bibinfo{booktitle}{2020 {USENIX} Annual Technical
  Conference ({USENIX} {ATC} 20)}}, \bibinfo{pages}{493--506}
  (\bibinfo{publisher}{{USENIX} Association}, \bibinfo{year}{2020}).

\bibitem{ma-mkckks}
\bibinfo{author}{Ma, J.}, \bibinfo{author}{Naas, S.-A.}, \bibinfo{author}{Sigg,
  S.} \& \bibinfo{author}{Lyu, X.}
\newblock \bibinfo{journal}{\bibinfo{title}{Privacy-preserving federated
  learning based on multi-key homomorphic encryption}}.
\newblock {\emph{\JournalTitle{International Journal of Intelligent Systems}}}
  (\bibinfo{year}{2022}).

\bibitem{stripelis2020:dvw}
\bibinfo{author}{Stripelis, D.} \& \bibinfo{author}{Ambite, J.~L.}
\newblock \bibinfo{journal}{\bibinfo{title}{Accelerating federated learning in
  heterogeneous data and computational environments}}.
\newblock {\emph{\JournalTitle{arXiv:2008.11281}}}  (\bibinfo{year}{2020}).

\bibitem{stripelis2022:corrupted}
\bibinfo{author}{Stripelis, D.}, \bibinfo{author}{Abram, M.} \&
  \bibinfo{author}{Ambite, J.~L.}
\newblock \bibinfo{journal}{\bibinfo{title}{Performance weighting for robust
  federated learning against corrupted sources}}.
\newblock {\emph{\JournalTitle{arXiv:2205.01184}}}  (\bibinfo{year}{2022}).

\bibitem{wang2021field}
\bibinfo{author}{Wang, J.} \emph{et~al.}
\newblock \bibinfo{journal}{\bibinfo{title}{A field guide to federated
  optimization}}.
\newblock {\emph{\JournalTitle{arXiv preprint arXiv:2107.06917}}}
  (\bibinfo{year}{2021}).

\bibitem{DBLP:conf/iclr/ReddiCZGRKKM21}
\bibinfo{author}{Reddi, S.~J.} \emph{et~al.}
\newblock \bibinfo{title}{Adaptive federated optimization}.
\newblock In \emph{\bibinfo{booktitle}{9th International Conference on Learning
  Representations, {ICLR} 2021, Virtual Event, Austria, May 3-7, 2021}}
  (\bibinfo{publisher}{OpenReview.net}, \bibinfo{year}{2021}).

\bibitem{hsu2019measuring}
\bibinfo{author}{Hsu, T.-M.~H.}, \bibinfo{author}{Qi, H.} \&
  \bibinfo{author}{Brown, M.}
\newblock \bibinfo{journal}{\bibinfo{title}{Measuring the effects of
  non-identical data distribution for federated visual classification}}.
\newblock {\emph{\JournalTitle{arXiv preprint arXiv:1909.06335}}}
  (\bibinfo{year}{2019}).

\bibitem{dwork2014algorithmic}
\bibinfo{author}{Dwork, C.} \& \bibinfo{author}{Roth, A.}
\newblock \bibinfo{journal}{\bibinfo{title}{{The Algorithmic Foundations of
  Differential Privacy}}}.
\newblock {\emph{\JournalTitle{Foundations and Trends in Theoretical Computer
  Science}}} \textbf{\bibinfo{volume}{9}}, \bibinfo{pages}{211--407},
  \doiprefix\url{10.1561/0400000042} (\bibinfo{year}{2014}).

\bibitem{jayaraman2019evaluating}
\bibinfo{author}{Jayaraman, B.} \& \bibinfo{author}{Evans, D.}
\newblock \bibinfo{title}{{Evaluating Differentially Private Machine Learning
  in Practice}}.
\newblock In \emph{\bibinfo{booktitle}{28th {USENIX} Security Symposium
  ({USENIX} Security 19)}}, \bibinfo{pages}{1895--1912}
  (\bibinfo{publisher}{{USENIX} Association}, \bibinfo{year}{2019}).

\bibitem{veitch2019understanding}
\bibinfo{author}{Veitch, D.~P.} \emph{et~al.}
\newblock \bibinfo{journal}{\bibinfo{title}{Understanding disease progression
  and improving alzheimer's disease clinical trials: Recent highlights from the
  alzheimer's disease neuroimaging initiative}}.
\newblock {\emph{\JournalTitle{Alzheimer's \& Dementia}}}
  \textbf{\bibinfo{volume}{15}}, \bibinfo{pages}{106--152}
  (\bibinfo{year}{2019}).

\end{thebibliography}

\newpage

\section*{Supplemental Material}

\renewcommand{\figurename}{Supplementary Figure}
\renewcommand{\thefigure}{S(\arabic{figure})}

\begin{figure*}[htpb]
\centering
  \subfloat[Centralized 100\%]{
    \includegraphics[width=0.4\linewidth]{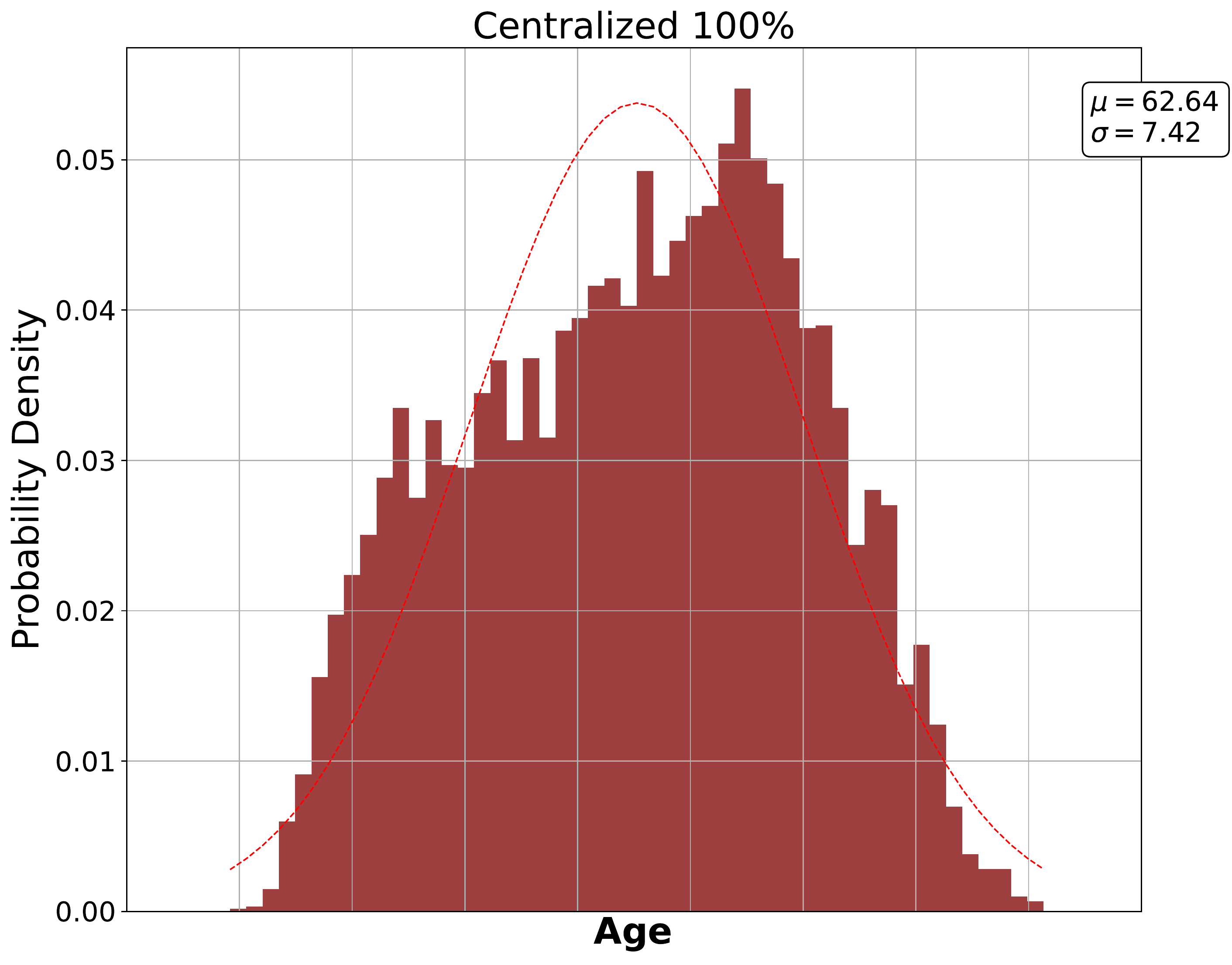}
    \label{subfig:UKBB_AgeDistributions_Centralized100pct}
  }
  \subfloat[Test Set]{
    \includegraphics[width=0.4\linewidth]{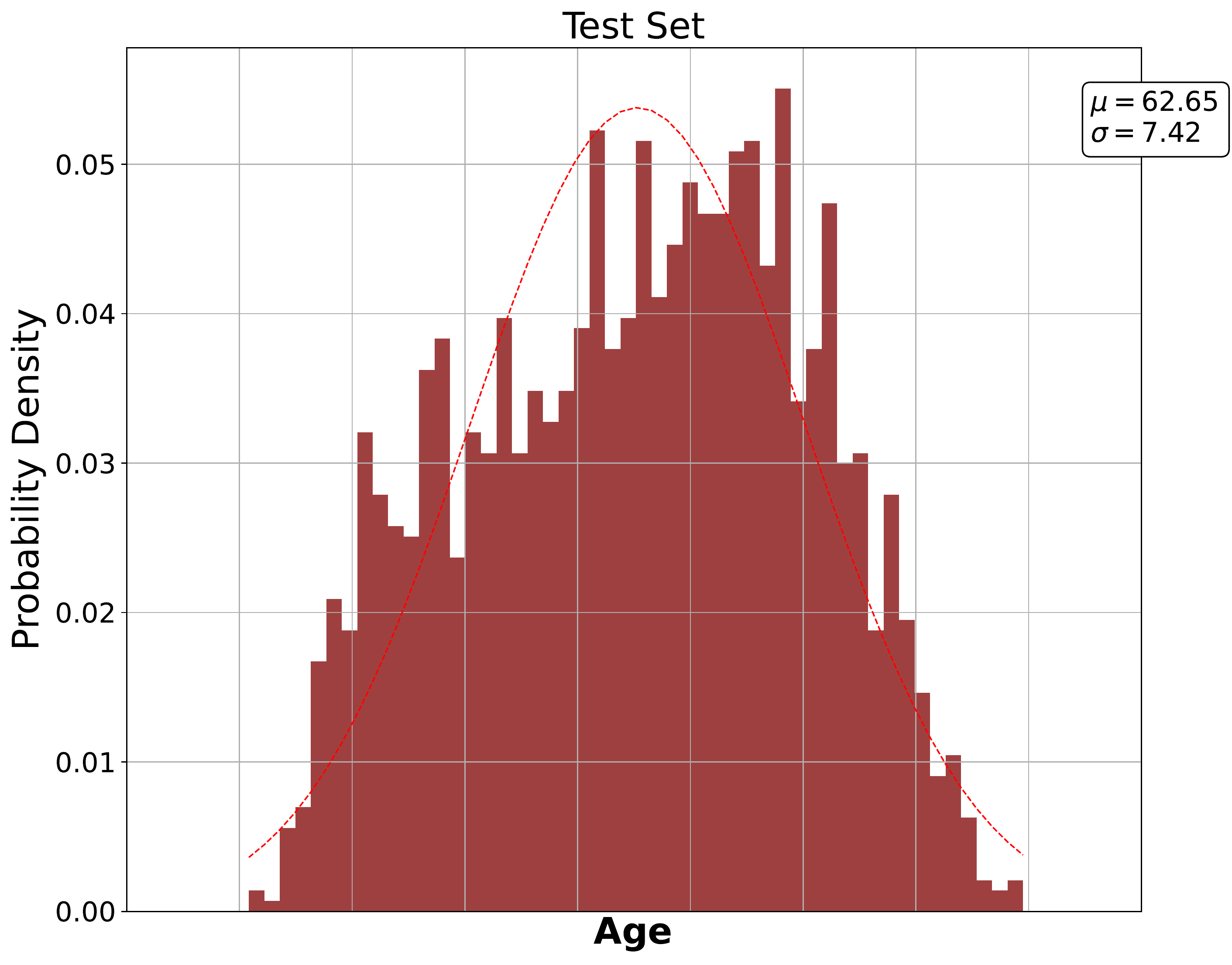}
    \label{subfig:UKBB_AgeDistributions_TestSet}
  }
    \caption{The data distribution of all UKBB MRI scans in the train and test set.}
  \label{fig:UKBB_Age_Distributions_ALL}
\end{figure*}

\begin{figure*}[htpb]
\centering
  \subfloat[Centralized 20\%]{
    \includegraphics[width=0.4\linewidth]{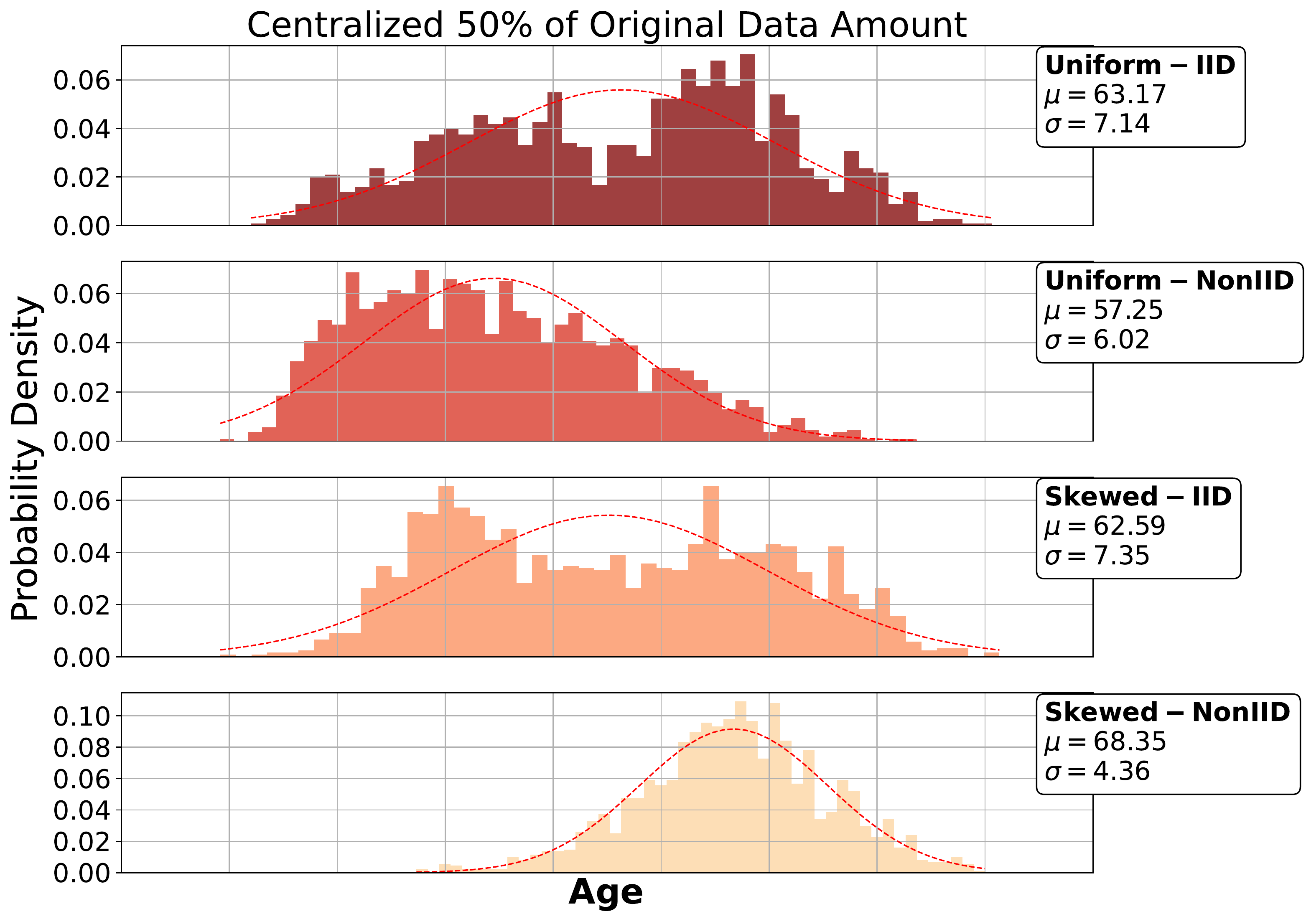}
    \label{subfig:UKBB_AgeDistributions_Centralized20pct}
  }
  \subfloat[Centralized 50\%]{
    \includegraphics[width=0.4\linewidth]{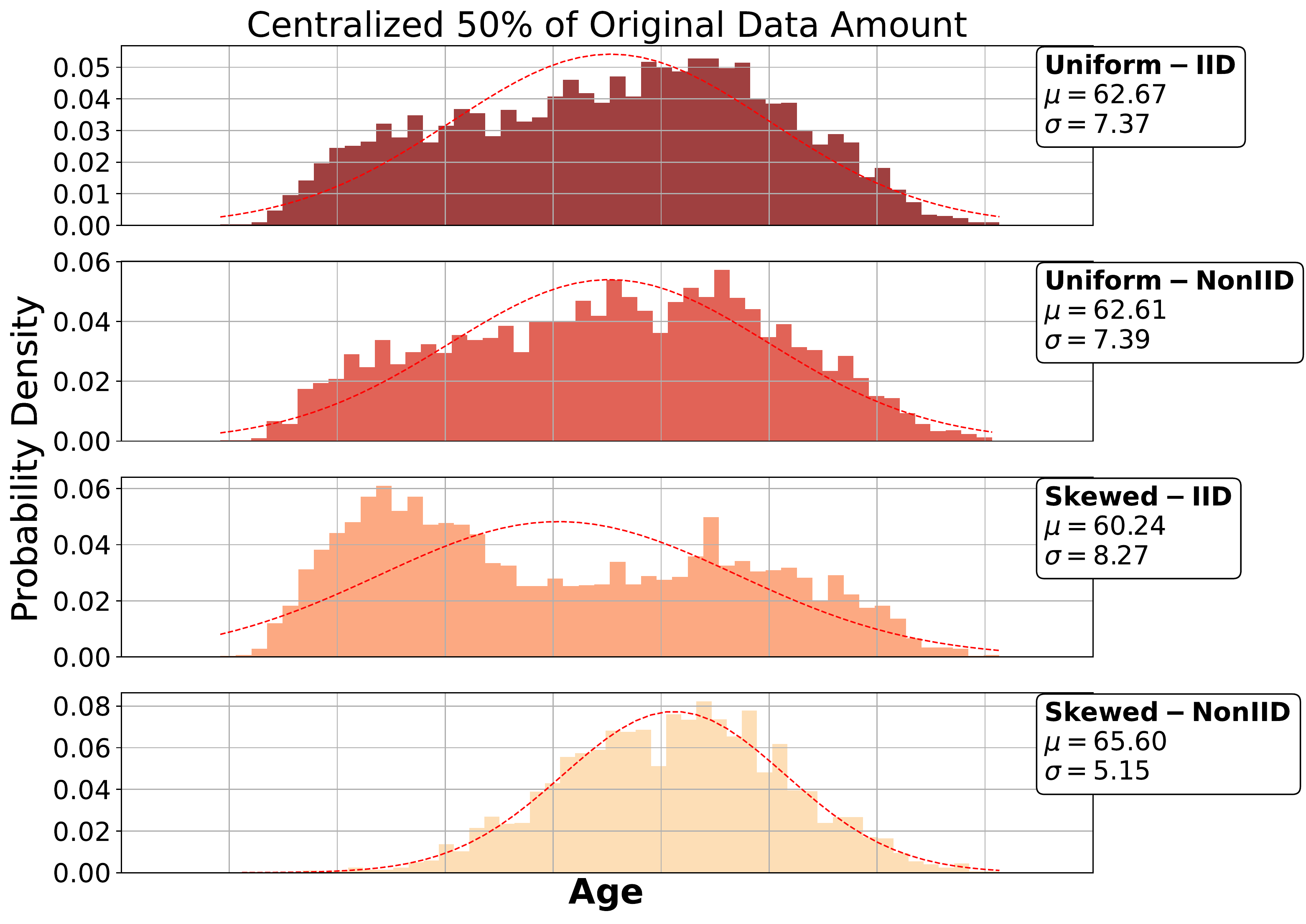}
    \label{subfig:UKBB_AgeDistributions_Centralized50pct}
  }
  \caption{Data distribution of the UKBB dataset for the statistically heterogeneous centralized learning environments, where the amount of data records accounts for the 20\% and 50\% of the original data samples.}
  \label{fig:UKBB_Centralized_Age_Distributions}
\end{figure*}

\begin{figure*}[htpb]
\centering
  \subfloat[Uniform \& IID]{
    \includegraphics[width=0.225\linewidth]{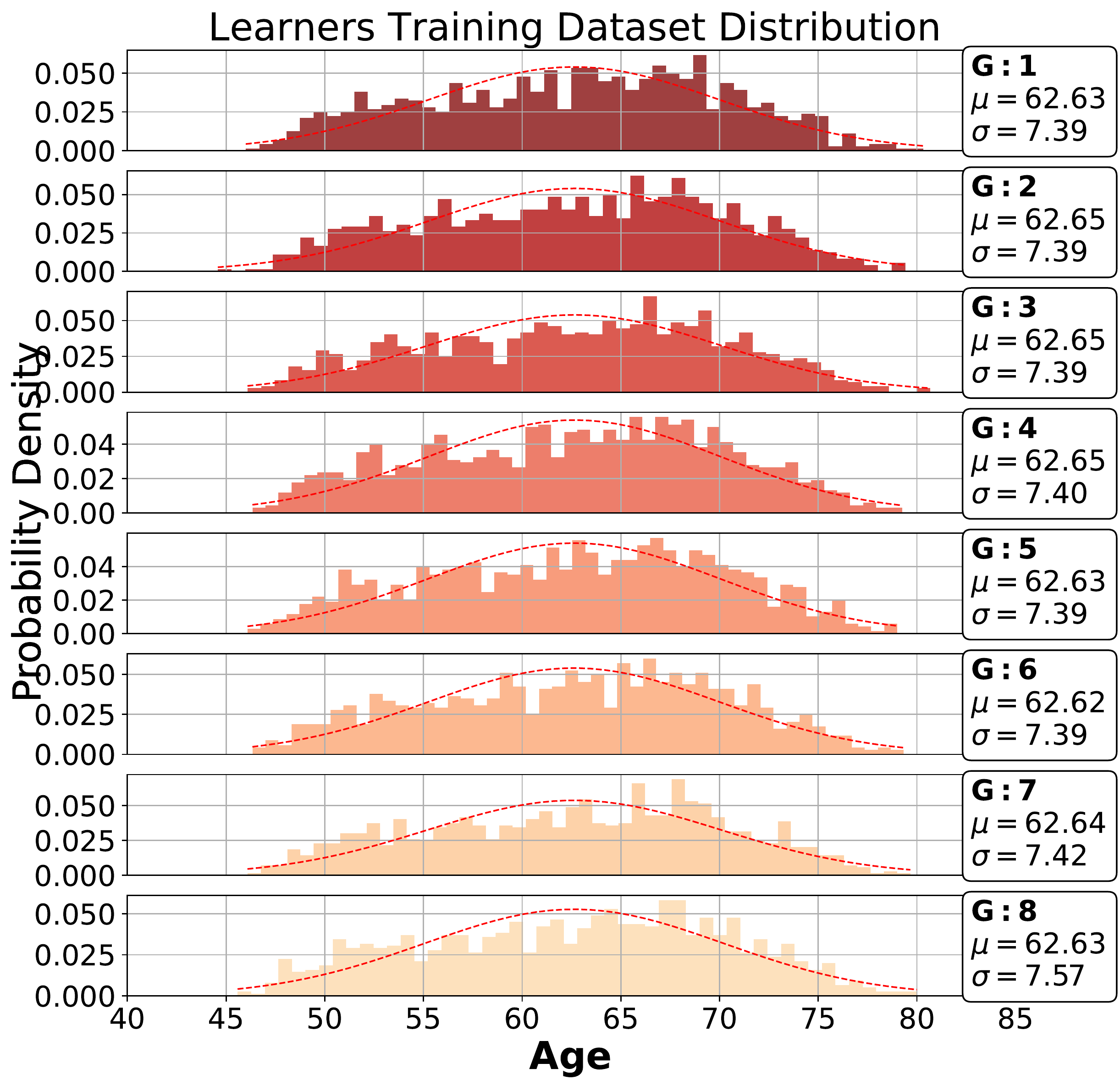}
    \label{subfig:UKBB_AgeDistributions_Uniform_IID}
  }
  \subfloat[Uniform \& Non-IID]{
    \includegraphics[width=0.225\linewidth]{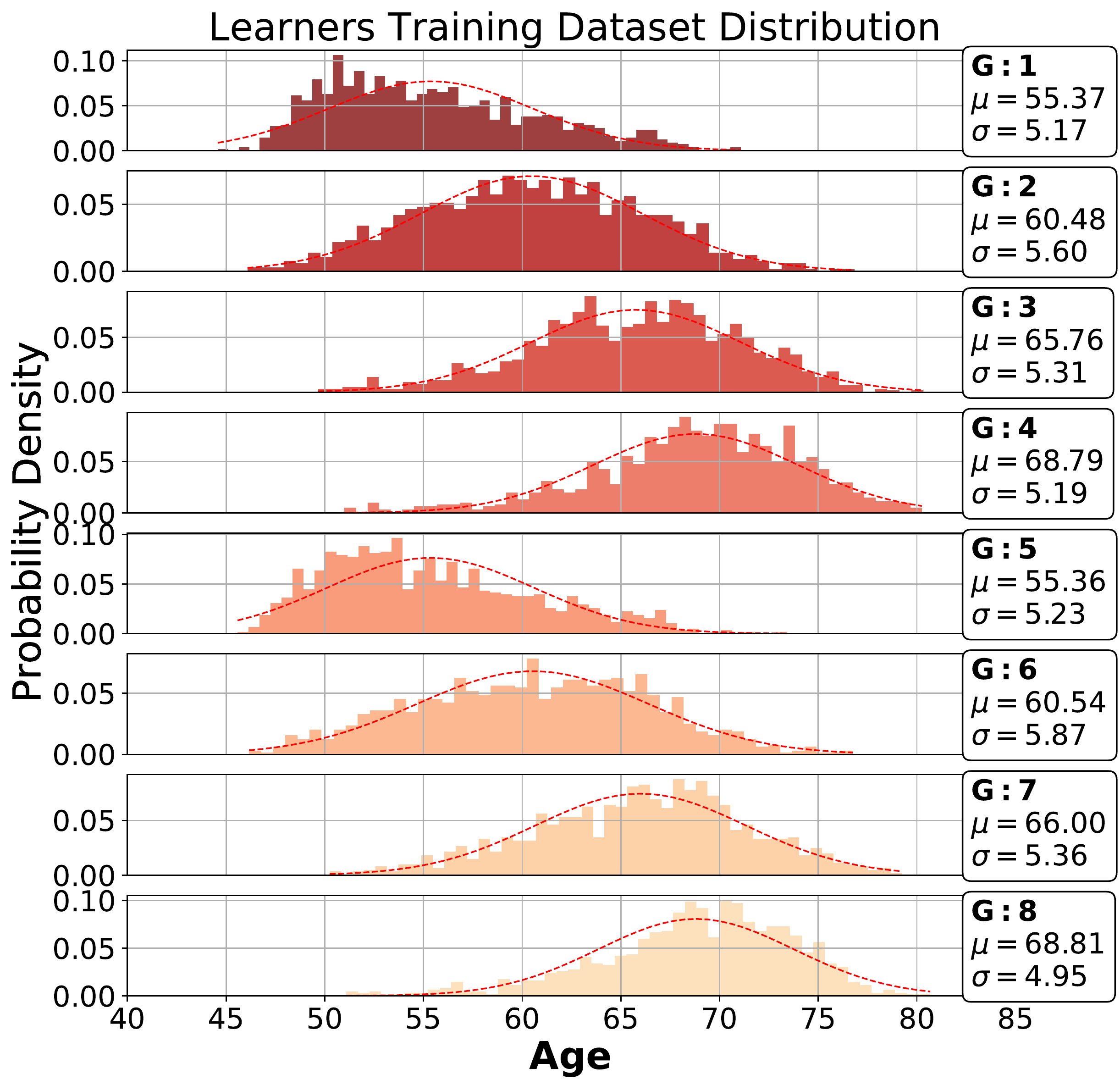}
    \label{subfig:UKBB_AgeDistributions_Uniform_NonIID}
  }
    \subfloat[Skewed \& IID]{
    \includegraphics[width=0.225\linewidth]{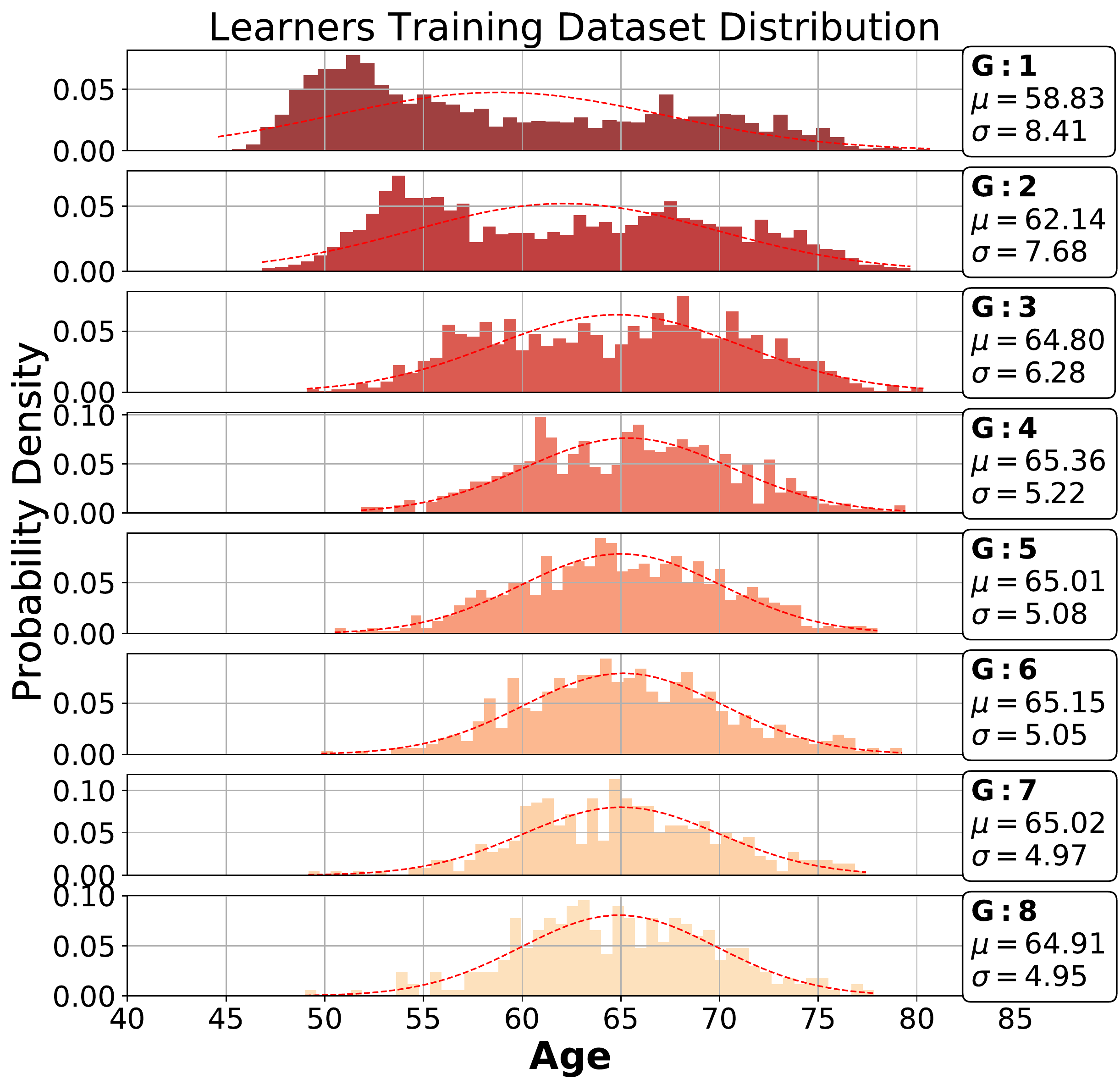}
    \label{subfig:UKBB_AgeDistributions_Skewed_IID}
  }
  \subfloat[Skewed \& Non-IID]{
\includegraphics[width=0.225\linewidth]{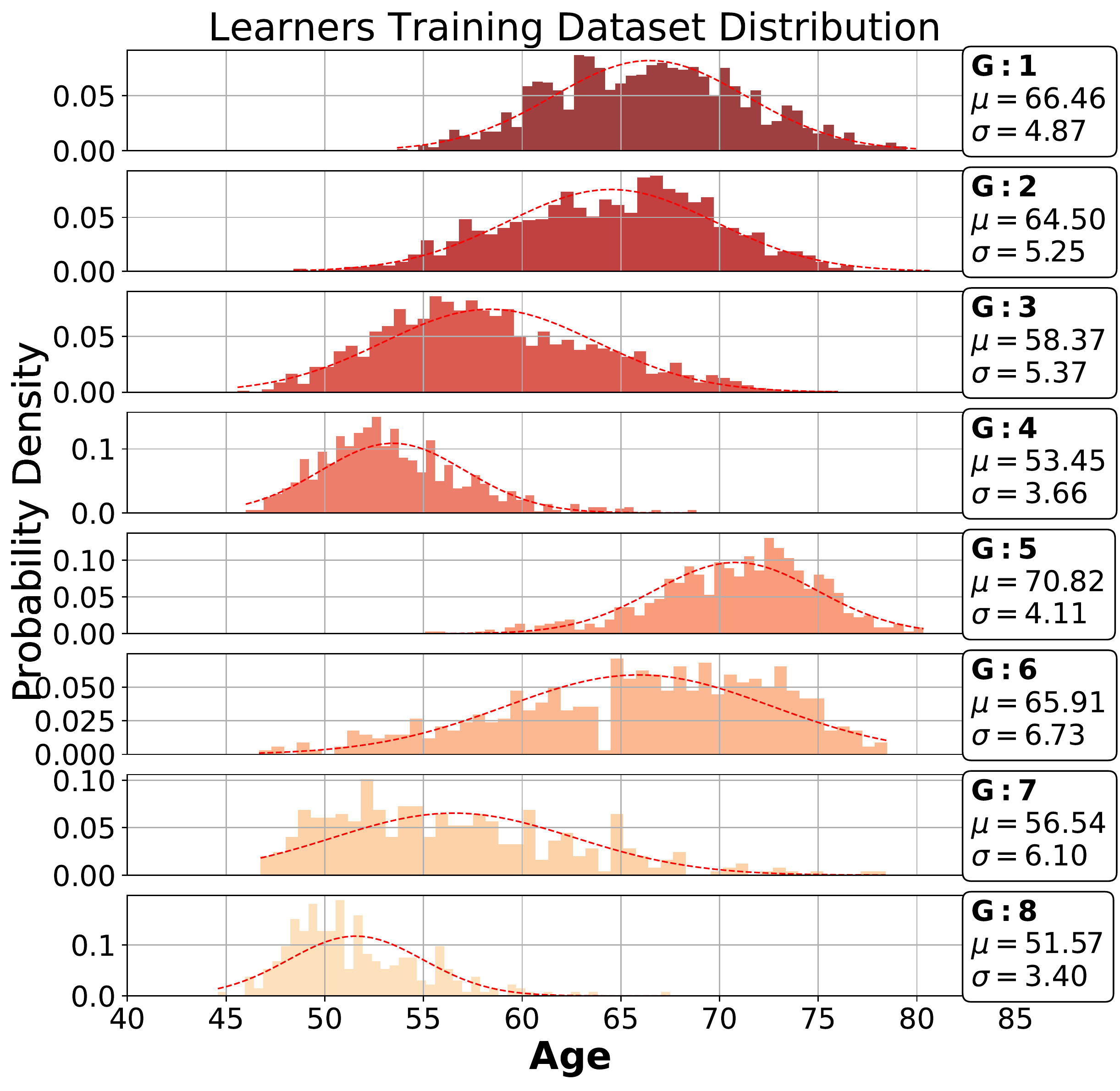}
    \label{subfig:UKBB_AgeDistributions_Skewed_NonIID}
  }
  \caption{The age distribution of the UKBB MRI scans allocated to each silo/learner (denoted as G:$i$) for the four federated learning environments investigated in this work.}
  \label{fig:UKBB_Federation_Age_Distributions}
\end{figure*}

\paragraph{UKBB Data Distributions.} Figure~\ref{fig:UKBB_Age_Distributions_ALL} presents the probability density of the chronological age associated with all the available UKBB MRI scans used for training and testing. We refer to the centralized model that trains over all available training MRI scans as Centralized-100\% (subfigure~\ref{subfig:UKBB_AgeDistributions_Centralized100pct}). The test set shown in~\ref{subfig:UKBB_AgeDistributions_TestSet} represents the common test set used to evaluate all models that may be trained on different data partitions or under different setup (e.g., either centralized or federated) of the UKBB dataset. The box to the right of each subplot shows each presented density's age mean and standard deviation.
Similarly, Figures~\ref{fig:UKBB_Centralized_Age_Distributions} and~\ref{fig:UKBB_Federation_Age_Distributions} present the data distribution of the MRI scans allocated for training the centralized-\{20\%, 50\%\} and federated models, respectively. 

\begin{figure}[htpb]
    \centering
    \hspace{25mm}
    \includegraphics[width=0.5\linewidth]{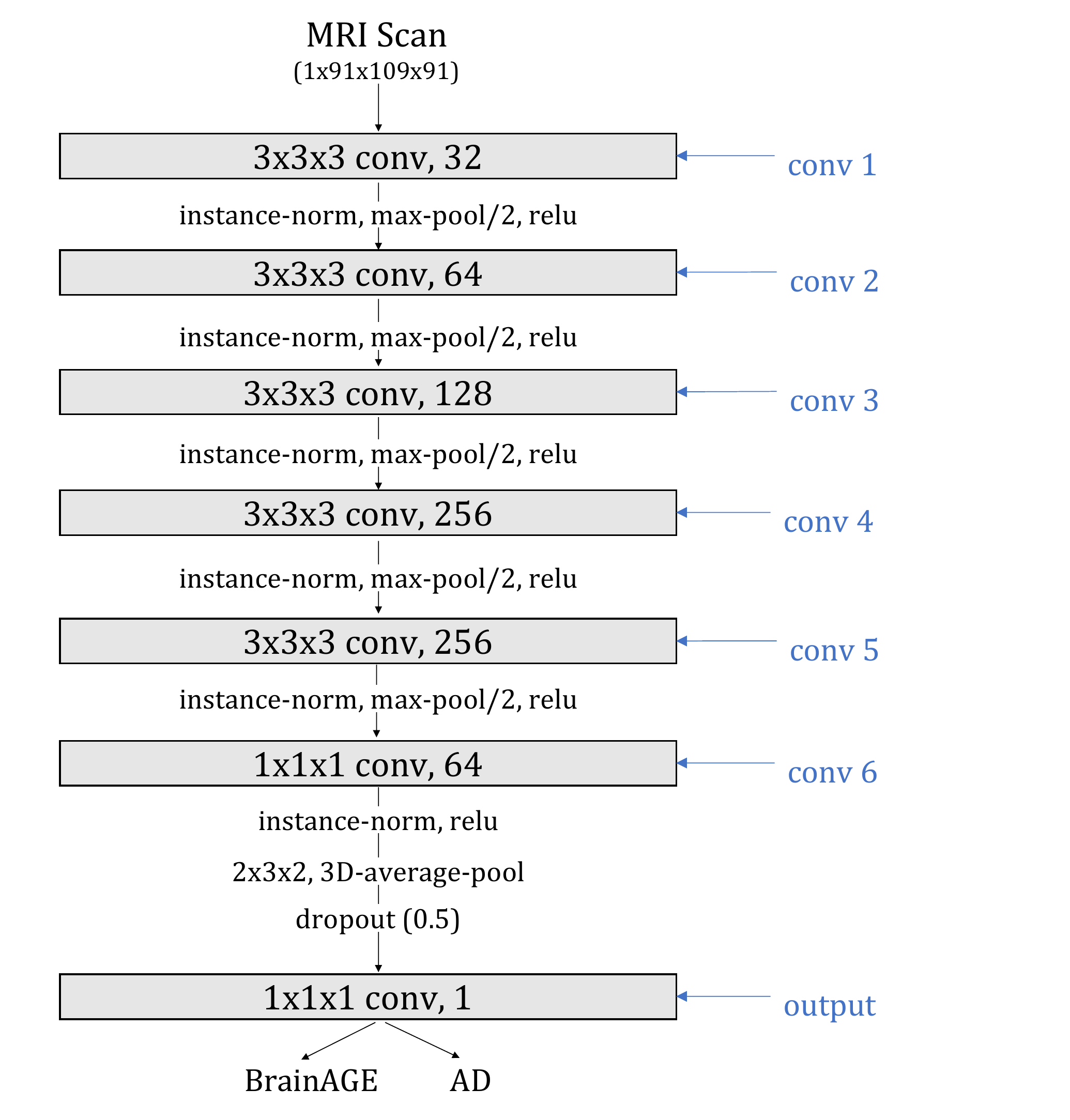}
    \caption{The deep learning model architecture used for the BrainAGE or AD prediction neuroimaging tasks. We train the models to do one task at a time. }
    \label{fig:3dcnn_model_definition}
\end{figure}

\paragraph{NN Architectures.} To perform our experiments for the BrainAGE and Alzheimer's Disease (AD) detection tasks, we used the same Convolutional Neural Network architecture. As shown in~\ref{fig:3dcnn_model_definition} our network consists of 6 convolution layers and one final prediction (output) layer. The final network layer is the only difference between the networks used to learn the BrainAGE and the AD tasks. For the BrainAGE regression task, we use a convolution layer of a single filter and kernel size, and for the AD binary classification task a dense layer of a single neuron. The total number of trainable network parameters is 2,950,401.

\paragraph{Federated Training Communication Cost.} We also analyze the communication cost incurred during federated training for all the federated training policies presented in this work. We measure the communication cost in terms of Gigabits based on the total number of model parameters exchanged during federated training. Specifically, during synchronous (and semi-synchronous) federated training, the total number of exchanged parameters equals $R * 2 * M$, where $R$ represents the total number of federation rounds, and $M$ is the total number of parameters. Factor $2$ accounts for the global model received by each learner at the beginning of training by the federation controller and for the local model sent by the learner to the controller at the end of training. In the case of asynchronous training, since no synchronization point exists among the learners, the communication cost is proportional to the total number of model update requests issued during training, i.e., $U * 2 * M$, where $U$ represents the total number of update requests. 

\begin{figure}[htpb]
    \centering
    \includegraphics[width=0.8\linewidth]{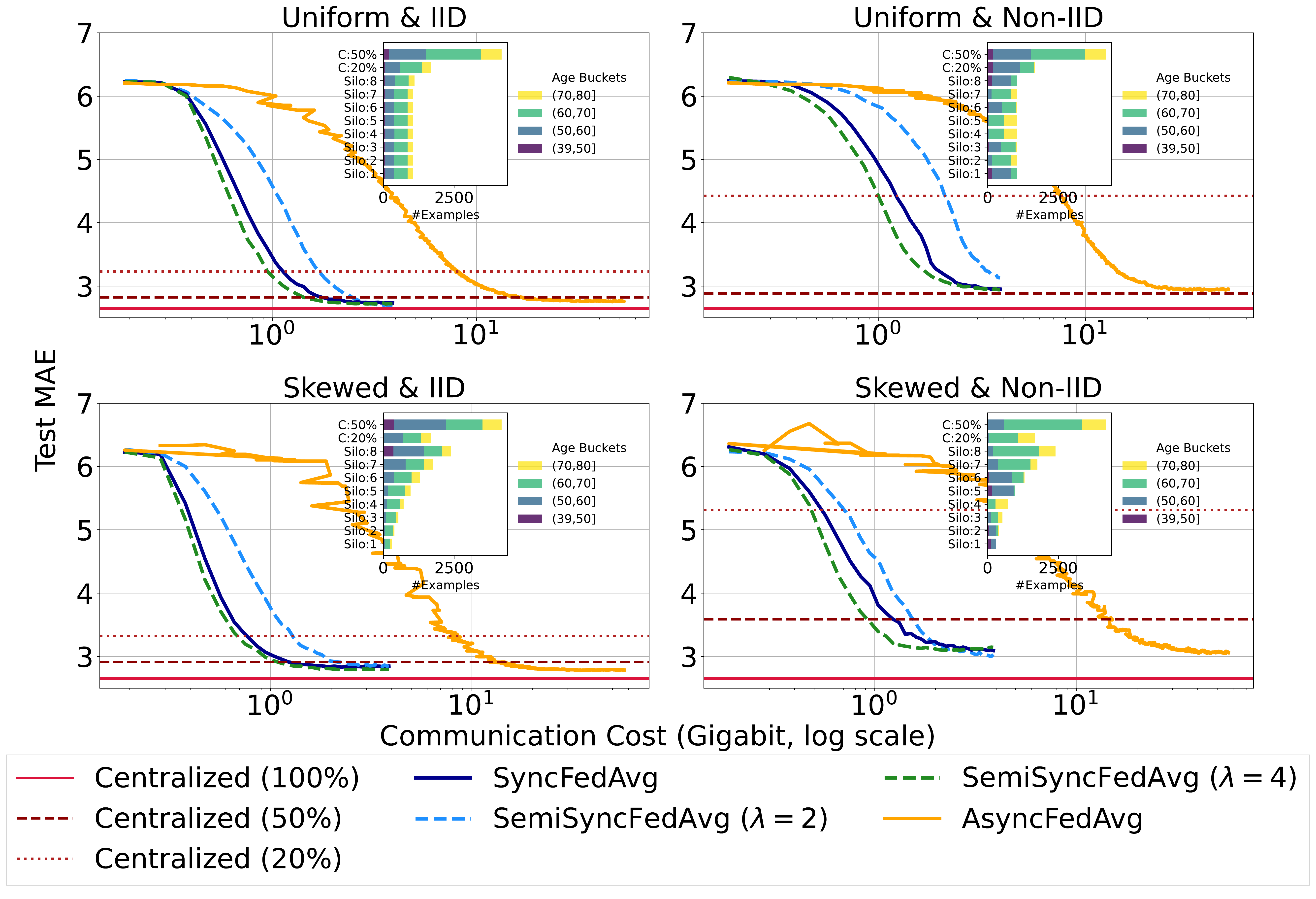}
    \caption{A comparison of synchronous, semi-synchronous, and asynchronous federated training policies based on communication cost.}
    \label{fig:BrainAge3D_PoliciesConvergence_CommunicationCost}
\end{figure}

The convergence of the federated models in terms of communication cost for the BrainAGE prediction task is shown in Figure~\ref{fig:BrainAge3D_PoliciesConvergence_CommunicationCost}. The figure shows that the asynchronous protocol (AsyncFedAvg) has competitive learning performance but requires significantly more communication cost (almost 10-times more). On the contrary, both SyncFedAvg and SemiSyncFedAvg require a lot less communication with SemiSync leading to convergence  with much lower communication cost when it is directly compared to SyncFedAvg. Relative to the differences in  communication cost, the time to convergence is very similar for all federated training methods (Figure ~\ref{fig:brainage_policies_convergence_wall_clock_time}).

\begin{figure}[htpb]
    \centering
    \includegraphics[width=0.8\linewidth]{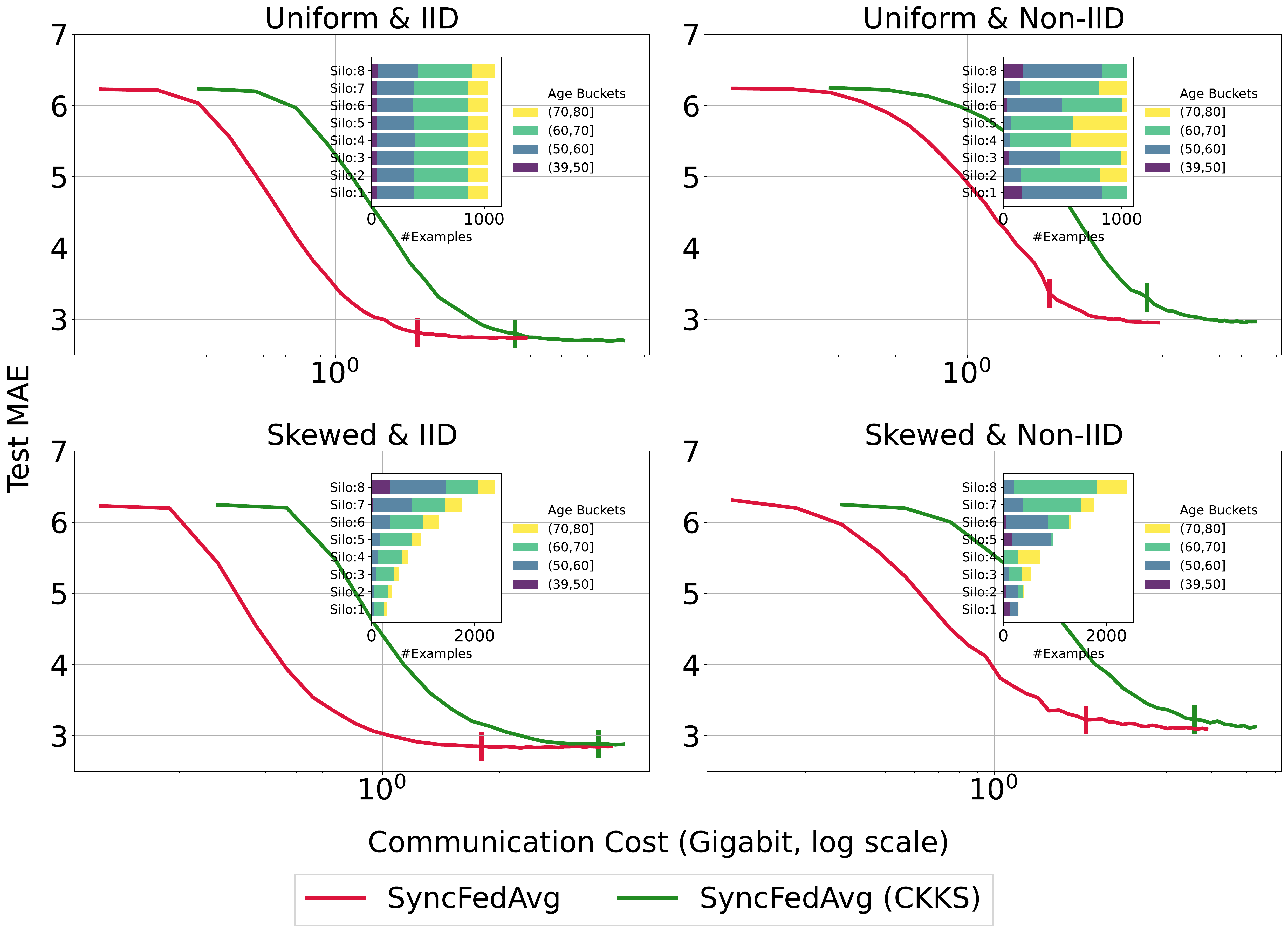}
    \caption{A comparison of the communication cost between synchronous federated average with (SyncFedAvg) and without encryption (SyncFedAvg (CKKS)).}
    \label{fig:BrainAge3D_FHE_PoliciesConvergence_CommunicationCost}
\end{figure}

Similarly, we analyze the communication cost when federated training is performed with and without encryption. As shown in Figure~\ref{fig:BrainAge3D_FHE_PoliciesConvergence_CommunicationCost} the communication cost incurred by the encrypted federated average approach is almost twice that of the non-encrypted approach. This communication overhead is due to the size of the encrypted model (ciphertext) exchanged during federated training. The ciphertext representation of the neural network requires the values to be represented as 64-bits which is twice the original bit size (32-bits) of the plaintext model used during the non-encrypted training.

\end{document}